\documentclass{article}

\usepackage{arxiv}

\usepackage[utf8]{inputenc} 
\usepackage[T1]{fontenc}    
\usepackage{hyperref}       
\usepackage{url}            
\usepackage{booktabs}       
\usepackage{amsfonts}       
\usepackage{nicefrac}       
\usepackage{microtype}      
\usepackage{lipsum}
\usepackage{bm} 
\usepackage{graphicx}
\graphicspath{ {./images/} }

\usepackage{natbib}
\usepackage{amsmath}
\usepackage{multirow}
\usepackage{subfigure} 
\usepackage{float}
\title{Binned semiparametric Bayesian networks for efficient kernel density estimation}

\author{
 Rafael Sojo \\
  Aingura IIoT, Paseo Mikeletegui 43, 20009 Donostia-San Sebastián, Gipuzkoa, Spain \\
  Universidad Politécnica de Madrid, Departamento de Inteligencia Artificial, 28660 Boadilla del Monte, Madrid, Spain \\
  \texttt{rsojo@ainguraiiot.com} 
   \And
 Javier Díaz-Rozo \\
  Aingura IIoT, Paseo Mikeletegui 43, 20009 Donostia-San Sebastián, Gipuzkoa, Spain \\
  \texttt{jdiaz@ainguraiiot.com} 
  \And
 Concha Bielza \\
  Universidad Politécnica de Madrid, Departamento de Inteligencia Artificial, 28660 Boadilla del Monte, Madrid, Spain \\
  \texttt{mcbielza@fi.upm.es} 
  \And
  Pedro Larrañaga \\
  Universidad Politécnica de Madrid, Departamento de Inteligencia Artificial, 28660 Boadilla del Monte, Madrid, Spain \\
  \texttt{pedro.larranaga@fi.upm.es} 
}

\begin{document}
\maketitle
\vspace{0.5cm}
\begin{abstract}
This paper introduces a new type of probabilistic semiparametric model that takes advantage of data binning to reduce the computational cost of kernel density estimation in nonparametric distributions. Two new conditional probability distributions are developed for the new binned semiparametric Bayesian networks, the sparse binned kernel density estimation and the Fourier kernel density estimation. These two probability distributions address the curse of dimensionality, which typically impacts binned models, by using sparse tensors and restricting the number of parent nodes in conditional probability calculations. To evaluate the proposal, we perform a complexity analysis and conduct several comparative experiments using synthetic data and datasets from the UCI Machine Learning repository. The experiments include different binning rules, parent restrictions, grid sizes, and number of instances to get a holistic view of the model's behavior. As a result, our binned semiparametric Bayesian networks achieve structural learning and log-likelihood estimations with no statistically significant differences compared to the semiparametric Bayesian networks, but at a much higher speed. Thus, the new binned semiparametric Bayesian networks prove to be a reliable and more efficient alternative to their non-binned counterparts. 
\end{abstract}


\section{Introduction}
Probabilistic graphical models (PGMs) are well-known tools for using graph-based representations to encode complex distributions over high-dimensional spaces \citep{koller2009probabilistic}. From the PGM family, Bayesian networks are one of the most popular types for factorizing the joint probability distribution (JPD) of a set of random variables. 
For continuous domains, there are two main types of Bayesian network models: parametric \citep{LGBNs} and nonparametric \citep{KDEBN}. 
Parametric models assume that the problem can be modeled using a known probability distribution with a finite number of parameters, while nonparametric models assume that the distribution is unknown and may involve a potentially infinite number of parameters. In other words, in nonparametric Bayesian networks, no specific assumption about the data distribution is made. However, a new type of Bayesian network was introduced recently in an attempt to reduce the time-complexity problems associated with nonparametric models. These are the semiparametric Bayesian networks (SPBNs) \citep{SPBN}, a model that combines the characteristics of both types. The parametric part of SPBNs involves Gaussian distributions, while the conditional probability distribution (CPD) of the non-Gaussian variables is calculated using the kernel density estimation (KDE) \citep{scott2015multivariate}. This method provides a much more flexible solution than a simple multivariate Gaussian distribution. However, the cost of the KDE algorithm is $O(N^2)$ for $N$ data points, making it computationally expensive for large sample sizes.

Several methods have been proposed to reduce this computational cost. For instance, \cite{silverman82} introduced a fast Fourier transform (FFT) approximation to accelerate the computation of univariate KDE models. Subsequently, \cite{Wand94} extended this approach to multivariate KDE using constrained (diagonal) bandwidth matrices, while \cite{gramacki2018} developed a solution that supports any symmetric and positive definite bandwidth matrix. Another notable contribution to FFT-based KDE (FKDE) models in univariate density estimation was presented by \cite{obrien2014}. Here, the authors reduced the cost of the empirical characteristic function of \cite{pigolotti2011} by a factor of 100. This approach was eventually generalized to multivariate settings by \cite{obrien2016}. However, to take advantage of the FFT, all these methods require a prior step of data discretization. In KDE, this process is commonly known as binning. The accuracy of binned KDE (BKDE) models has been extensively studied in the literature \citep{jones89, hall1996accuracy, pawlak99} and is widely acknowledged as an effective method to reduce the computational cost of KDEs. 
To control the accuracy, \cite{raykar_fast_2010} proposed a fast algorithm that reduces the computational complexity of univariate KDE through a Taylor series expansion for the Gaussian kernel. This expansion allows for an efficient $\epsilon$-exact approximation. Similarly, \cite{TangKarunamuni2016} proposed the refined linear binning and centered binning methods for univariate and multivariate KDE, which allow exact computations of the kernel density. In contrast to \cite{raykar_fast_2010}, \cite{TangKarunamuni2016} require the use of the symmetric beta family of kernel functions.

Although there is extensive literature on BKDEs, the challenges posed by the curse of dimensionality limit their applicability to tackle high-dimensional problems. For instance, in conjunction with Bayesian networks to accelerate the factorization of JPDs. Specifically, the curse refers to the exponential growth in computational demands as the number $n$ of variables increases, since both BKDEs and FKDEs rely on the construction of $n$-dimensional grids. Neither of the current multivariate KDE approximations reviewed has been evaluated in more than 3 dimensions.
For that reason, we propose a solution that enables the integration of BKDE and FKDE models with SPBNs to work with a higher number of variables. Thus, we introduce a novel binned SPBN (B-SPBN) that performs KDE operations faster than a standard SPBN while maintaining minimal structural and log-likelihood differences in a trade-off between speed and precision. 

The paper is organized as follows. Section \ref{sec:back} provides an overview of the fundamental concepts of Bayesian networks and SPBNs. Section \ref{sec:binning} introduces the univariate and multivariate data binning. Section \ref{sec:bsbns} presents the new B-SPBNs. Section \ref{sec:experiments} discusses the experimental results. Section \ref{sec:conclusion} concludes the paper and provides potential future research directions.

\section{Bayesian networks}
\label{sec:back}
A Bayesian network is a machine learning model denoted as $\mathcal{B} = (\mathcal{G}, \bm{\theta})$ that comprises a direct acyclic graph (DAG) $\mathcal{G}$ and a set of parameters $\bm{\theta}$, for which $\mathcal{G} = (V, A)$ is defined as a set of nodes $V$ and a set of arcs $A \subseteq V \times V$ between these nodes. For each $\mathcal{B}$ we have a dataset $\mathcal{D}$ with $n$ different variables in a vector $\mathbf{X} = (X_1, ..., X_n)$ and $N$ instances, i.e., $\mathcal{D} = \{\textbf{x}^{1},...,\textbf{x}^{N} \}$.  
Each arc of $A$ in $\mathcal{G}$ represents a probabilistic dependence between variables, for instance, in $X_1 \xrightarrow{} X_2$, $X_1$ is referred to as the parent of $X_2$. Therefore, for each DAG there is a set of conditional dependences and independences that define the parameters $\bm{\theta}$ of the CPD associated to each variable. Then, the JPD of the network can be factorized using these CPDs. For continuous variables, each CPD can be seen as a conditional probability density function (PDF) such that the JPD factorizes as: 
\begin{equation}
    f(\textbf{x}) = \prod_{i=1}^n\ f(x_i|\textbf{x}_{\text{Pa}(i)}) \ ,
    \label{eqn:jpd_pdf}
\end{equation}
where $\text{Pa}(i)$ denotes the parents of $X_i$. 

\subsection{Semiparametric Bayesian networks}
SPBNs integrate both parametric and nonparametric CPDs, adapting their behavior to the problem's demands. For parametric CPDs, they use linear Gaussian (LG) CPDs \citep{LGBNs}:

\begin{equation}
    f_{\text{LG}}(x_i|\textbf{x}_{\text{Pa}(i)}) = \mathcal{N}(\beta_{i0} + \sum_{k \in \text{Pa}(i)} \beta_{ik} x_{k}, \sigma_i^2) \ ,
    \label{eqn:lg_cpd}
\end{equation}
where $\beta_{i0}$ is the intercept associated to node $i$ in the regression of all parents of $X_i$ over $X_i$ and $\beta_{ik}$ is the coefficient associated to parent $k$. Note that $\sigma_i^2$ is the variance and does not depend on $\text{Pa}(i)$.
In contrast, nonparametric CPDs are defined using conditional KDE (CKDE) CPDs \citep{KDEBN}. The density function of a KDE model in the univariate case is:
\begin{equation}
    f_{\text{KDE}}(x) = \frac{1}{N} \sum_{j=1}^N K_{h}(x-x^{j})  \ ,
    \label{eqn:univariate_kde}
\end{equation}
where $x^j$ denotes the $j$-th training instance, $h$ represents the bandwidth parameter, and 
$K_{h}(x-x^{j}) = \frac{1}{h}K(\frac{x-x^{j}}{h})$ refers to the scaled univariate kernel function. 
The generalization to a multivariate KDE is straightforward: 
\begin{equation}
    f_{\text{KDE}}(\textbf{x}) = \frac{1}{N} \sum_{j=1}^N K_{\textbf{H}}(\textbf{x}-\textbf{x}^{j}) \ ,
    \label{eqn:multivariate_kde}
\end{equation}
where $\textbf{x}^{j} = (x_1^j,\dots , x_n^j)$, $\textbf{H}$ is the $n\times n$ symmetric and positive definite bandwidth matrix and $K_\textbf{H}(\textbf{x}-\textbf{x}^{j}) = |\textbf{H}|^{-1/2} K( |\textbf{H}|^{-1/2} (\textbf{x}-\textbf{x}^{j}))$ denotes the scaled multivariate kernel function. There are several kernel functions \citep{Chacon2018,Heidenreich2013} such as the Gaussian kernel, the Epanechnikov kernel or the biweight kernel.  

\begin{figure}[!ht]
\centering
\includegraphics[width=0.25\linewidth]{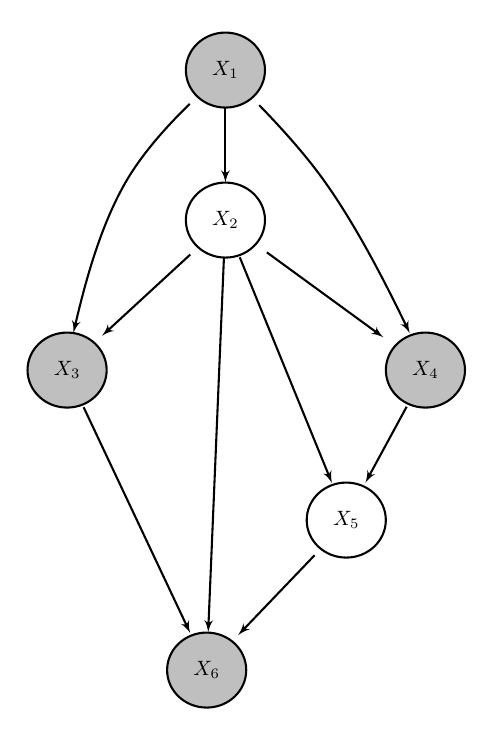}
\caption{Example of a SPBN structure.}
\label{fig:example_spbn}
\end{figure}

For the calculation of conditional probabilities, let $f_{\text{KDE}}(x_i,\textbf{x}_{\text{Pa}(i)})$ denote the joint KDE model for $X_i$ and $\textbf{X}_{\text{Pa}(i)}$ and $f_{\text{KDE}}(\textbf{x}_{\text{Pa}(i)})$ the marginal KDE model for $\textbf{X}_{\text{Pa}(i)}$. By using the Bayes' theorem, the conditional distribution of $X_i$ given $\textbf{X}_{\text{Pa}(i)}$ in a CKDE CPD is:

\begin{equation}
    f_{\text{CKDE}}(x_i|\textbf{x}_{\text{Pa}(i)}) = \frac{f_{\text{KDE}}(x_i,\textbf{x}_{\text{Pa}(i)})}{f_{\text{KDE}}(\textbf{x}_{\text{Pa}(i)})} =
    \frac{\sum_{j=1}^N K_{\textbf{H}_i}\left(
    \begin{bmatrix} 
    x_i\\
    \textbf{x}_{\text{Pa}(i)}\\
    \end{bmatrix} - 
    \begin{bmatrix} 
    x_{i}^{j}\\ 
    \textbf{x}_{\text{Pa}(i)}^{j}\\ 
    \end{bmatrix}\right)}{\sum_{j=1}^N K_{\textbf{H}_{i}^{-}}
    (\textbf{x}_{\text{Pa}(i)} - \textbf{x}_{\text{Pa}(i)}^{j})} \ ,
    \label{eq:non-parm-cpd}
\end{equation}
where $\mathbf{H}_i$ and $\mathbf{H}_{i}^{-}$ are the joint and marginal bandwidth matrices for $f_{\text{KDE}}(x_i,\textbf{x}_{\text{Pa}(i)})$ and $f_{\text{KDE}}(\textbf{x}_{\text{Pa}(i)})$, respectively. Figure \ref{fig:example_spbn} illustrates an example of a SPBN structure, with nodes using LG CPDs shown in white and nodes using CKDE CPDs in gray.

\subsection{Parameter learning}
\label{sec:parameter learning}
To learn the parameters that compose the CPDs of a SPBN there are different approaches.
In the Gaussian estimation, the parameters of the CPDs are $\beta_{i0}$, $\beta_{ik}$ with $k\in \text{Pa}(i)$ and $\sigma_i^2$, for which the standard maximum likelihood estimate is used. Thus, assuming independent identically distributed samples in a dataset $\mathcal{D}$, the log-likelihood $\mathcal{L}$ of $\mathcal{D}$ given a DAG $\mathcal{G}$ and some parameters $\bm{\theta}$ is:

\begin{equation}
    \mathcal{L}(\bm{\theta}) = \mathcal{L}( \mathcal{D} | \mathcal{G},\bm{\theta}) = \sum_{j=1}^N \sum_{i=1}^n\ \log f(x_{i}^{j}|\textbf{x}_{\text{Pa}(i)}^{j})
    \label{eq:logl}
\end{equation}

Then, the parameters $\bm{\theta}$ that maximize $\mathcal{L}(\bm{\theta})$ are estimated such that:
\begin{equation}
    \hat{\bm{\theta}}^{\text{MLE}} = arg \max_{\bm{\theta} \in \Theta} \mathcal{L}(\bm{\theta})
\end{equation}

On the other hand, the parameters of a CKDE node $X_i$ are the joint and marginal bandwidth matrices $\mathbf{H}_i$ and $\mathbf{H}_{i}^{-}$, for which we will use the normal reference rule extended to the multivariate case \citep{normal-reference-rule}:
\begin{equation}
    \hat{\mathbf{H}}_i = \left(\frac{4}{n+2}\right)^{2/(n+4)} \hat{\Sigma} N^{-2/(n+4)} \ ,
\end{equation}
where $\hat{\Sigma}$ is the sample covariance matrix of $X_i$ and $\textbf{X}_{\text{Pa}(i)}$. 
The normal reference rule provides a closed-form solution that minimizes the asymptotic approximation of the mean integrated squared error (AMISE) in multivariate density estimation. The AMISE is an approximation to the mean integrated squared error (MISE) when $N \xrightarrow{} \infty$. The MISE is given by the expectancy of the integrated squared error (ISE). That is:
\begin{equation}
    \text{MISE}\{\hat{f}\} = \mathbb{E} \left[ \text{ISE}\{\hat{f}\} \right] = \mathbb{E} \left[ \int_{\mathbb{R}^n} \left(\hat{f}(x) - f(x)\right)^2 dx\right]
\end{equation}

Usually, minimizing the AMISE can only be performed numerically, but for normal mixture densities it can be computed analytically. Consequently, the optimal bandwidth matrix can be determined.

\section{Data binning}
\label{sec:binning}
As mentioned above, a KDE model for a dataset $\mathcal{D}$, according to Equation (\ref{eqn:multivariate_kde}),  has a complexity of $O(N^2)$ for $N$ data points. However, in most practical applications it is more efficient to compute the KDE for equally spaced points. This approach is commonly known as binning, a way of grouping continuous values into smaller number of bins. For the computation of the KDE, binning can be understood as a kind of data discretization. Depending on whether this is done for one or multiple dimensions, the binning process involves different considerations. 

\subsection{Univariate binning}
\label{sec:univaraite_binning}
Let $M$ denote the size of a grid consisting of equally spaced, ordered points  $\{g^{1},...,g^{M} \}$, $g^1 < \dots < g^M$, with corresponding weights  $\{c^{1},...,c^{M}\}$. Depending on its value, each sample point $x$ can be assigned to a grid point $g^m$, along with its associated weight $c^m$. 
\begin{equation}
   x \xrightarrow{} \{g^{m},c^{m}\},\ m = 1,\dots , M 
\end{equation}

\begin{figure}[H]
    \centering
    \includegraphics[width=0.75\linewidth]{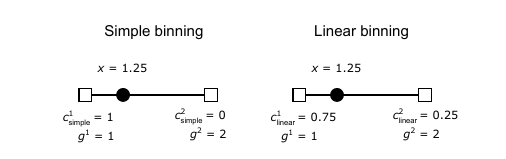}
    \caption{Univariate data binning ($\delta=1$).}
    \label{fig:1d_binning}
\end{figure}

Weights are calculated based on neighboring observations, for which the most common binning procedures are the simple and linear binning rules. For the simple binning rule, a weight of 1 is assigned to the closest grid point $g^m$ of every sample $x$, while in the linear binning rule this weight is spread over the surrounding grid points, where closer grid points weight more (see Figure \ref{fig:1d_binning}). Take for instance $x \in [g^{m}, g^{m+1}]$, the simple binning rule can be defined as:
\begin{equation}
    c_{\text{simple}}^{m} =
    \begin{cases}
      1 & \text{if}\  (x-g^{m}) < (g^{m+1}-x) \\
      0 & \text{otherwise}
    \end{cases}
    \label{eq:simple_bin}
\end{equation}

For the linear binning rule, the weights associated to the surrounding grid points are:
\begin{align}
    c_{\text{linear}}^{m} &= \frac{ g^{m+1} - x }{\delta} \ , \\
    c_{\text{linear}}^{m+1} &= \frac{ x - g^{m}}{\delta} \ ,
\end{align}
where $\delta = (g^{M}- g^{1})/(M-1)$ is the grid binwidth and in both cases (simple and linear) $\sum_{m=1}^{M} c^{m} = N$. Note that the weights of multiple sample points placed into the same grid point $g^m$ are added together. 

\subsection{Multivariate binning}
\begin{figure}[!h]
    \centering
    \includegraphics[width=1\linewidth]{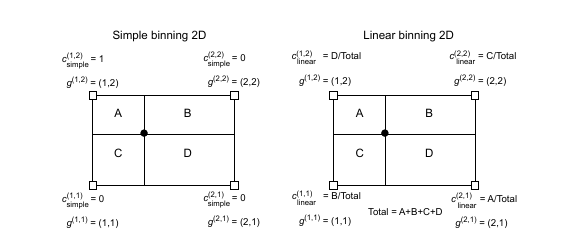}
    \caption{Bivariate data binning.}
    \label{fig:2d_binning}
\end{figure}

In the multivariate case, let $M_i$ denote the size of a grid for dimension $i$, consisting of equally spaced, ordered points $\{g_i^{1},...,g_i^{M_i}\}$, $g_i^{1} < \dots < g_i^{M_i}$ with corresponding weights $\{c_i^{1},...,c_i^{M_i}\}$. As in the previous section, any sample vector $\textbf{x}$ can be replaced by the grid vector $\textbf{g}^{\textbf{m}} = (g_1^{m_1}, \dots, g_n^{m_n})$, indexed by $\textbf{m} = (m_1,\dots, m_n)$, along with its corresponding weight $c^{\textbf{m}}$. However, in this case, the weight is determined by the product of univariate rules. Therefore, let $c^{\textbf{m}}$ be the weight corresponding to $\textbf{g}^{\textbf{m}}$, drawn from the weight tensor \textbf{C} of size $M_1 \times \dots \times M_n$. The weight $c^{\textbf{m}}$ can be computed as follows:
\begin{equation}
    c^{\textbf{m}} = \prod_{i=1}^{n} c_i^{m_i}
    \hspace{0.3cm}
    \text{with}
    \hspace{0.3cm}
    \sum_{\textbf{m} \in \textbf{M}} c^{\textbf{m}} = N \ ,
\end{equation}
where \textbf{M} denotes the Cartesian product of the sets of indices $\{1,\dots ,M_1\}  \times \dots \times \{1,\dots ,M_n\}$ with cardinal $M_1 \times \dots \times M_n$.
Figure \ref{fig:2d_binning} illustrates the simple and linear binning rules for the bivariate case. A, B, C and D are the areas of the corresponding rectangles. As described by \cite{Wand94}, in bivariate linear binning the contribution of \textbf{x} is distributed among the surrounding grid points based on areas of opposite subrectangles. For higher-dimensional data, these areas are replaced by volumes in both binning procedures. As noted earlier, this is equivalent to the product of univariate rules. 

\section{Binned semiparametric Bayesian networks}
\label{sec:bsbns}
In this section, we will present the new B-SPBNs. Our proposal takes advantage of data binning to accelerate the estimation of CPDs in non-Gaussian variables. In particular, we will use the BKDE and FKDE models described in \cite{gramacki2018}.
To mitigate the curse of dimensionality, the BKDE model will be implemented using sparse tensors. However, the FKDE will require a restriction in the number of parent nodes to prevent from memory overflow. Then, a brief complexity analysis will shed light to such limitations.

\subsection{Sparse binned kernel density estimation}
\label{sec:sbkde}
Considering the data binning process of Section \ref{sec:univaraite_binning}, Equation (\ref{eqn:univariate_kde}) can be rewritten as: 
\begin{equation}
    \hat{f}_{\text{BKDE}}(x) =  \frac{1}{N} \sum_{m=1}^{M} K_{h}(x-g^{m})c^{m}
    \label{eq:bkde_1d}
\end{equation}
where $\hat{f}$ is used instead of $f$ to denote that the BKDE is an approximation of the standard KDE. This change reduces the computational cost from $O(N^2)$ to $O(NM)$, which is more convenient since usually $M << N$.
For the multivariate case, Equation (\ref{eq:bkde_1d}) can be adapted to iterate through the binned space. That is:
\begin{equation}
    \hat{f}_{\text{BKDE}}(\textbf{x}) =  \frac{1}{N} \sum_{\textbf{m} \in \textbf{M}}K_{\textbf{H}}(\textbf{x}-\textbf{g}^{\textbf{m}}) c^{\textbf{m}}
    \label{eq:bkde_2d}
\end{equation}

Now, the cost of the model becomes $O(N M_1\cdots M_n)$.
Since the computational cost of a multivariate BKDE, as defined by Equation (\ref{eq:bkde_2d}), grows exponentially with the number of variables, it can become problematic when dealing with high-dimensional data. Nevertheless, this issue can be addressed by using sparse tensors to avoid processing weights that are equal to zero. A sparse tensor is a data structure where the number of non-zero elements is considerably smaller than the total number of them. This property is often exploited to save memory and computational resources by only storing and processing non-zero elements and their positions. 
Thus, building on Equation(\ref{eq:bkde_2d}), let $\hat{f}_{\text{SBKDE}}(\textbf{x})$ denote a sparse binned KDE (SBKDE) model as follows:
\begin{equation}
    \hat{f}_{\text{SBKDE}}(\textbf{x}) =  \frac{1}{N} \sum_{\textbf{s} \in \textbf{S}} K_{\textbf{H}}(\textbf{x}-\textbf{g}^{\textbf{s}})c^{\textbf{s}}_{+} \ ,
    \label{eq:sbkde}
\end{equation}
where $\textbf{S}$ is the Cartesian product of the sets of indices $\{1, \dots, M_1\} \times \dots \times \{1, \dots, M_n\}$, restricted to positions where $c^{\textbf{m}} > 0$. 
Let $S$ be the number of these positions.
Thus, $c^{\textbf{s}}_{+}$ denotes the weight associated to the grid vector $\textbf{g}^{\textbf{s}}$, indexed by $\textbf{s} = (s_1, \dots , s_n)$ with $s_i = 1,\dots , M_i$, from the sparse weight tensor $\textbf{C}_{+}$ of size $S$. Note that $S$ corresponds to the total number of non-zero weights from $\textbf{C}$. 

Thus, a new type of nonparametric CPD can be presented, the SBKDE distribution. 

\textbf{Definition}. Let $X_i$ be a  random variable following an SBKDE CPD. The conditional distribution of $X_i$ given $\textbf{X}_{\text{Pa}(i)}$ is defined as:

\begin{equation} 
    \hat{f}_{\text{SBKDE}}(x_i|\textbf{x}_{\text{Pa}(i)}) = \frac{\hat{f}_{\text{SBKDE}}(x_i, \textbf{x}_{\text{Pa}(i)} )}{\hat{f}_{\text{SBKDE}}(\textbf{x}_{\text{Pa}(i)} )} \ ,
    \label{eq:cases_SBKDE}
\end{equation}
where $\hat{f}_{\text{SBKDE}}(x_i, \textbf{x}_{\text{Pa}(i)} )$ and $\hat{f}_{\text{SBKDE}}(\textbf{x}_{\text{Pa}(i)} )$ are SBKDE models as in Equation (\ref{eq:sbkde}).

\subsection{Fourier binned kernel density estimation}
To construct the FKDE model, we can work on Equation (\ref{eq:bkde_1d}) to take the form of a convolution. Further details about the reformulation are given in \cite{gramacki2018}.

\subsubsection{Univariate case}
 For the univariate case, the FKDE model is defined as:
\begin{align}
    \hat{f}_{\text{FKDE}}(g^{t}) &= 
     \sum_{l =-L}^{L} 
     c^{t-l} k^{l} = (\textbf{c} \ast \textbf{k})^{t} \ ,  \label{eq:ftkde_univariate} 
     \\ 
     k^{l} &= \frac{1}{N} K_{h}(g^{l}) = \frac{1}{N} K_{h}(\delta l) \ , \nonumber
      \\
      L &= \min\left\{M-1,\lceil \frac{4 h}{\delta} \rceil \right\} \ , \nonumber
\end{align}
where $g^t$, $t=1,\dots ,M$, is a grid point, $\ast$ is the convolution operator,  \textbf{c} is the vector of weights and \textbf{k} is the vector of kernel values $k^l$. To ensure that both \textbf{c} and \textbf{k} have the same length, the \textit{zero-padding} procedure outlined by Gramacki is employed. Thus, the new length of the vectors is determined by:
\begin{equation}
 P= 2^{\lceil \log_{2}(3M-1) \rceil} \ ,
 \label{eq:size_P}
\end{equation}
where $\lceil \cdot \rceil$ is the ceiling operator. 
In this context, we can leverage the convolution theorem \citep{convTheo}, which states that a convolution  in time  domain is equivalent to a multiplication in the frequency domain. In other words, a point-wise product of Fourier transforms. Let $\mathcal{F}$ be the FFT operator \citep{1965-cooley}, $\mathcal{F}^{-1}$ the inverse, and $\textbf{c}_{zp}$ and $\textbf{k}_{zp}$ the \textit{zero-padding} vectors of size $P$. Accordingly, Equation (\ref{eq:ftkde_univariate}) can be solved such that:
\begin{equation}
    \hat{f}_{\text{FKDE}}(g^{t}) = (\textbf{c}_{zp} \ast \textbf{k}_{zp})^t = \mathcal{F}^{-1}\{\mathcal{F}(\textbf{c}_{zp}) \cdot \mathcal{F}(\textbf{k}_{zp})\}^{2M-1 + t} \ ,
    \label{eq:convolved}
\end{equation}
where $2M-1$ denotes the offset at which the densities are located after performing $\mathcal{F}^{-1}$.

\subsubsection{Multivariate case}
As in the previous subsection (Section \ref{sec:sbkde}), the generalization to a multivariate scenario requires iterating through the binned space. Hence:
\begin{align}
     \hat{f}_{\text{FKDE}}(\textbf{g}^{\textbf{t}}) &= 
     \sum_{\textbf{l} \in \textbf{L}} 
     c^{\textbf{t-l}} k^{\textbf{l}} = (\textbf{C} \ast \textbf{K})^{\textbf{t}} \ , \label{eq:ftkde_multivariate}
     \\ 
     k^{\textbf{l}} &= \frac{1}{N} K_{\textbf{H}}(\textbf{g}^{\textbf{l}}) = \frac{1}{N} K_{\textbf{H}}(\delta_1 l_1,\dots , \delta_n l_n) \ , \nonumber \\ 
    L_i &= \min\left\{M_i-1,\lceil \frac{4 \sqrt{|\lambda|}}{\delta_i} \rceil \right\} \ , \nonumber
\end{align}
where $\textbf{L}$ denotes the Cartesian product of the sets of indices $\{-L_1,\dots ,L_1\}  \times \dots \times \{-L_n,\dots ,L_n\}$ with cardinal $L_1 \times \dots \times L_n$, $|\lambda|$ corresponds to the largest absolute eigenvalue of $\textbf{H}$ and $\textbf{g}^{\textbf{t}} = (g_1^{t_1}, \dots, g_n^{t_n})$, with $\textbf{t} = (t_1,\dots, t_n)$. Likewise, let $\textbf{C}_{zp}$ and $\textbf{K}_{zp}$ denote two \textit{zero-padding} tensors with size $P_1 \times \dots \times P_n$, $P_i= 2^{\lceil \log_{2}(3M_i-1) \rceil}$. The convolution can be solved such that:
\begin{equation}
    \hat{f}_{\text{FKDE}}(\textbf{g}^{\textbf{t}}) = (\textbf{C}_{zp} \ast \textbf{K}_{zp})^\textbf{t} = \mathcal{F}^{-1}\{\mathcal{F}(\textbf{C}_{zp}) \cdot \mathcal{F}(\textbf{K}_{zp})\}^{\textbf{d}} \ ,
    \label{eq:convolved_multivariate}
\end{equation}
where $\textbf{d}=(2M_i-1+t_1, \dots, 2M_n-1+t_n)$.

Now, the FKDE CPDs can be presented. 

\textbf{Definition}. Let $G_i$ be a binned random variable following an FKDE CPD. The conditional distribution of $G_i$ given $\textbf{G}_{\text{Pa}(i)}$ is defined as:

\begin{equation} 
     \hat{f}_{\text{FKDE}}(g_i|\textbf{g}_{\text{Pa}(i)}) = \frac{\hat{f}_{\text{FKDE}}(g_i, \textbf{g}_{\text{Pa}(i)} )}{\hat{f}_{\text{FKDE}}(\textbf{g}_{\text{Pa}(i)})} \ ,
    \label{eq:cases_fkde}
\end{equation}
where $\hat{f}_{\text{FKDE}}(g_i, \textbf{g}_{\text{Pa}(i)} )$ and $\hat{f}_{\text{FKDE}}(\textbf{g}_{\text{Pa}(i)} )$ are FKDE models as in Equation (\ref{eq:convolved_multivariate}).

\subsection{Complexity analysis}
\label{sec:computational_cost} 
The B-SPBN aims to be a computationally efficient alternative to the SPBN. As we introduced previously, this is a more flexible type of continuous Bayesian network that combines both parametric and nonparametric computations. For that reason, B-SPBNs share all the theoretical properties of the standard SPBNs and are thought to be used in high-dimensional datasets, for which the factorization of the JPD allows more accurate and efficient density estimations. There are multiple approaches to reduce the cost of KDEs, but as we reviewed in the introduction, their scalability is restricted due to the curse of dimensionality. Some of these methods provide exact solutions for specific kernel functions, such as the Gaussian in \citep{raykar_fast_2010} or the symmetric beta in \citep{TangKarunamuni2016}. In our case, we propose the SBKDE and FKDE CPDs to approximate the CKDE CPDs. These methods are not exact, but allow for more flexibility, as they can be used with any kernel function and any symmetric and positive definite bandwidth matrix. In addition, their implementation within a continuous Bayesian network basis simplifies the task of handling high-dimensional spaces. In the next paragraphs, we will analyze their computational complexity and restrictions.

Beginning with the SBKDE, it is evident that the complexity of Equation (\ref{eq:sbkde}) is $O(NS)$, where the value of $S$ depends on the binning procedure. For the simple binning rule, the weight of each data point is assigned to the closest grid point; therefore, $S \leq N$. In contrast, linear binning distributes the weights across the surrounding grid points. As a result, \( 2^n \) weight values are computed for each data point. In both cases, weights falling into the same grid point are summed. According to \cite{hall1996accuracy}, linear binning requires fewer grid points to achieve the same 1\% relative mean integrated squared error as simple binning.
The use of sparse tensors alleviates the memory demands for storing $n$-dimensional grids, so we can argue that there is no memory constraint for either binning rule in most applications. However, even with small grid sizes, the number of weight values can grow significantly in high-dimensional spaces for the linear binning case.
In practice, this can result in higher computational complexities than the standard KDE.

\begin{figure}[!h]
    \centering
    \includegraphics[width=0.75\linewidth]{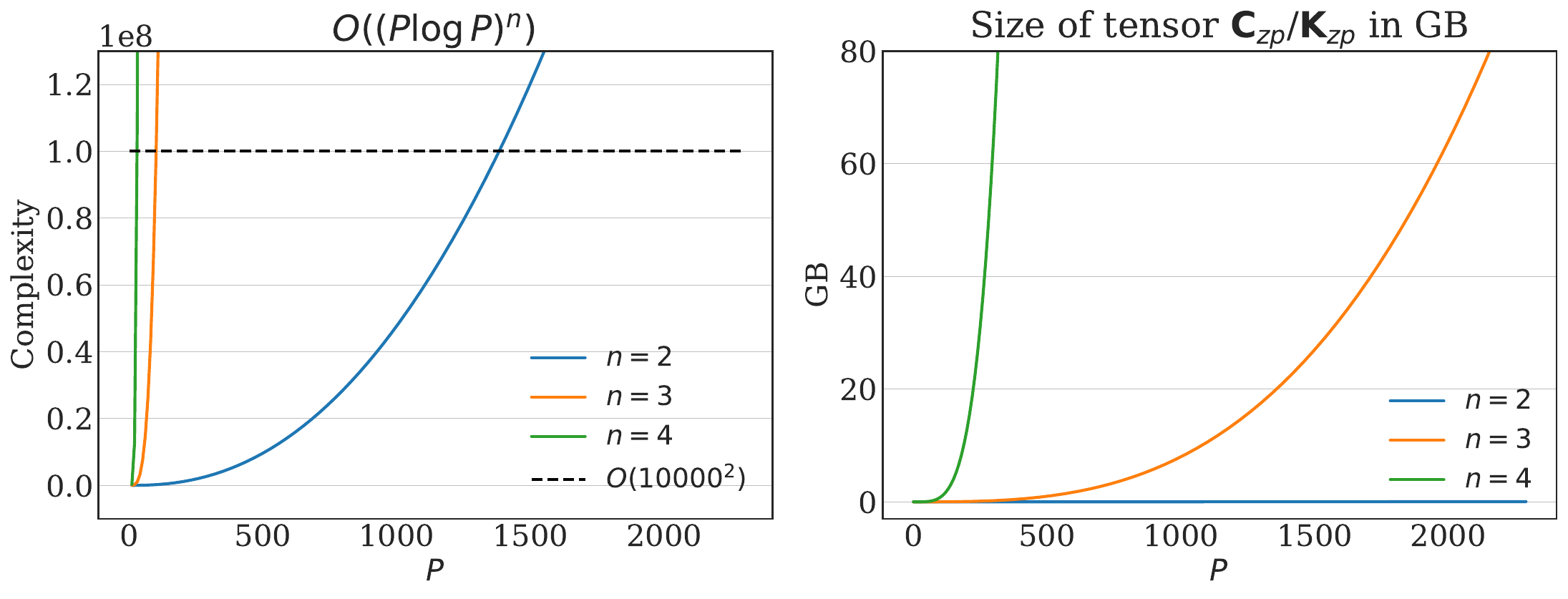}
    \caption{Growth of computational demands for FKDE CPD.}
    \label{fig:demands-fkde}
\end{figure}

On the other hand, the cost of the FKDE is $O(P_1\log P_1 \cdots P_n\log P_n)$, which is more efficient than $O(N^2)$ or $O(NS)$ in low-dimensional settings. Nevertheless, it requires the construction of two $n$-dimensional tensors of size $P_1 \times \dots \times P_n$, $\textbf{C}_{zp}$ and $\textbf{K}_{zp}$, that could become too large to fit in memory. 
For a better understanding of the problem, Figure \ref{fig:demands-fkde} illustrates the growth in computational demands for 2, 3, and 4 variables, considering the same $P$ value along all dimensions. The $x$-axis corresponds to the size of $P$ in both plots, while the $y$-axis is the time complexity (left) and the memory space (right) required to store an $n$-dimensional tensor of type double. To provide some context, the time complexity chart includes a dashed line indicating the cost of a KDE model with 10000 data points.
Note that in both charts, the curve grows exponentially as $P$ increases, with a notably steeper slope as $n$ becomes larger. Although $n=3$ may still fit in memory for a wide range of sizes of $P$, the time complexity escalates dramatically at relatively small values of $P$. Therefore, although the best performance is ensured with 1 or 2 dimensions, FKDE CPDs can still be used for 3 or 4 dimensions (2 or 3 parent nodes in the joint FKDE, Equation (\ref{eq:cases_fkde})) with smaller grid sizes before having any memory overflow or excessive execution times.

\section{Experimental results}
\label{sec:experiments}
In this section, we will evaluate the performance of the new B-SPBNs, using data sampled from synthetic functions (see Appendix \ref{apendix:synthetic_data}) and data from the UCI Machine Learning repository \citep{uci_repo}. 
The baseline for comparison in all the experiments is the standard SPBN, although we also employ GBNs in the experiments with data from the UCI Machine Learning repository to highlight the differences between parametric and semiparametric models. Since we aim to provide a faster and more accurate alternative to the SPBNs, we will focus our evaluation on the field of continuous Bayesian networks. Further details are given in the following subsections.

All networks will be learned using the same grid size $M$ for all dimensions and the largest $L_i$ among them. Additionally, we will use the Gaussian kernel $K(\textbf{x}) = (2\pi^{n/2})^{-1}e^{-\frac{1}{2}\textbf{x}^\text{T}\textbf{x}}$ for the CKDE, SBKDE, and FKDE CPDs.
The Gaussian kernel is a common kernel function because it leverages properties of Gaussian densities, like the fast calculation of marginal and conditional distributions or the infinite differentiability. Moreover, a KDE with this kernel is equivalent to a mixture model with each component located on each training instance. However, any other kernel with a valid $\textbf{H}$ could be used \citep{Chacon2018,kernel_smooth}.
To estimate the structures, we will use the greedy hill-climbing (HC) algorithm with a patience $\lambda = 3$ and the $k$-fold cross-validated log-likelihood score with 5 folds  \citep{SPBN}. HC is a score-based methodology adapted from \cite{hc} to deal with SPBNs. It is an optimization algorithm that moves over the space of DAGs, performing small changes to improve the score, such as arc additions, arc deletions, arc flips, and changes in the type of node. This algorithm produces approximate solutions that may vary between runs on the same dataset. For that reason, the structural learning experiments will be repeated 5 times each. 
To perform the experiments, we will use a modified version of the PyBNesian\footnote{ \url{https://repo.hca.bsc.es/gitlab/aingura-public/pybnesian}} library that executes in CPU.

\subsection{Synthetic datasets}
\label{sec:synthetic_datasets}

\begin{figure}[!h]
    \centering
    \subfigure[Structure of synthetic SPBN 1.]{
        \includegraphics[width=0.18\linewidth]{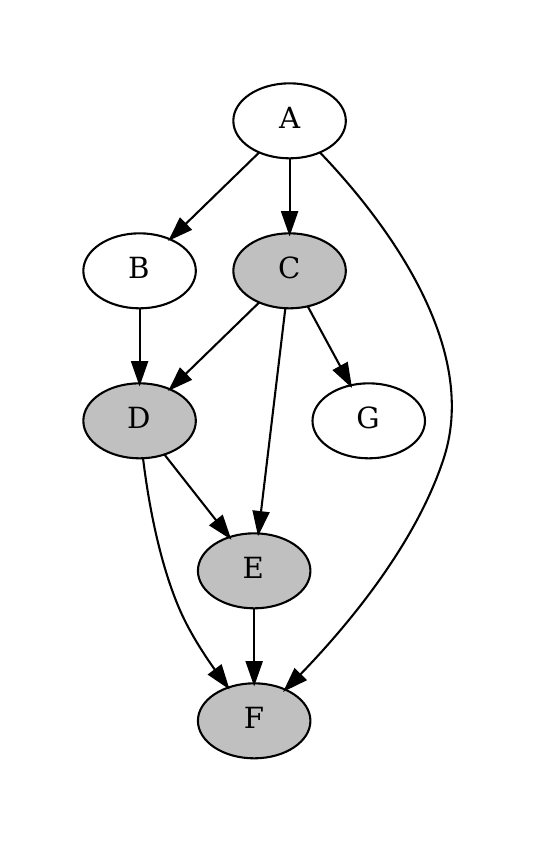}
        \label{fig:synthetic1}
    }
    \hfill
    \subfigure[Structure of synthetic SPBN 2.]{
        \includegraphics[width=0.29\linewidth]{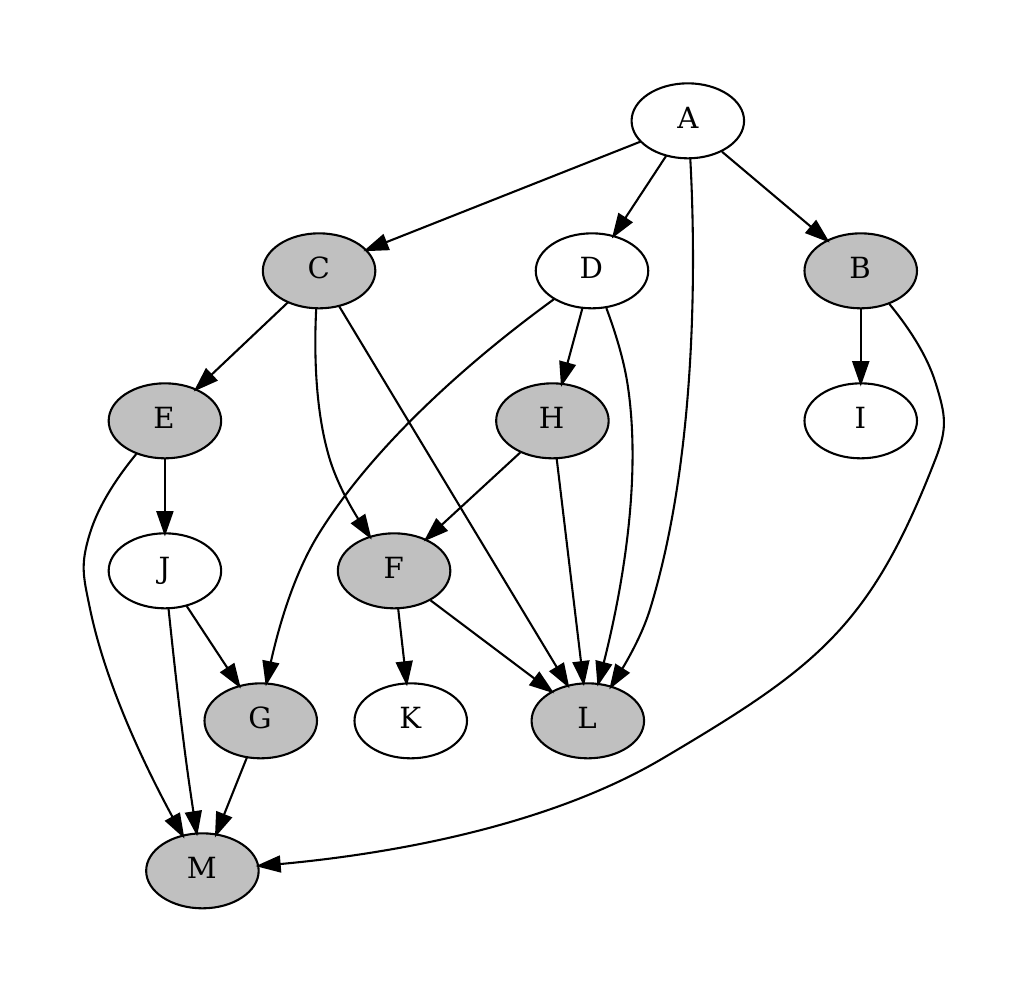}
        \label{fig:synthetic2}
    }
    \hfill
    \subfigure[Structure of synthetic SPBN 3.]{
        \includegraphics[width=0.24\linewidth]{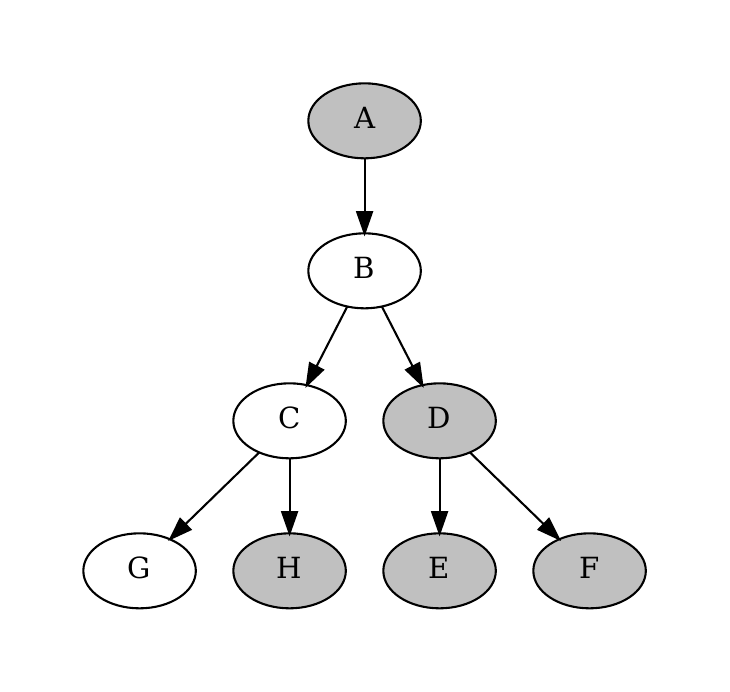}
        \label{fig:synthetic3}
    }    
    \hfill
    \subfigure[Structure of synthetic SPBN 4.]{
        \includegraphics[width=0.29\linewidth]{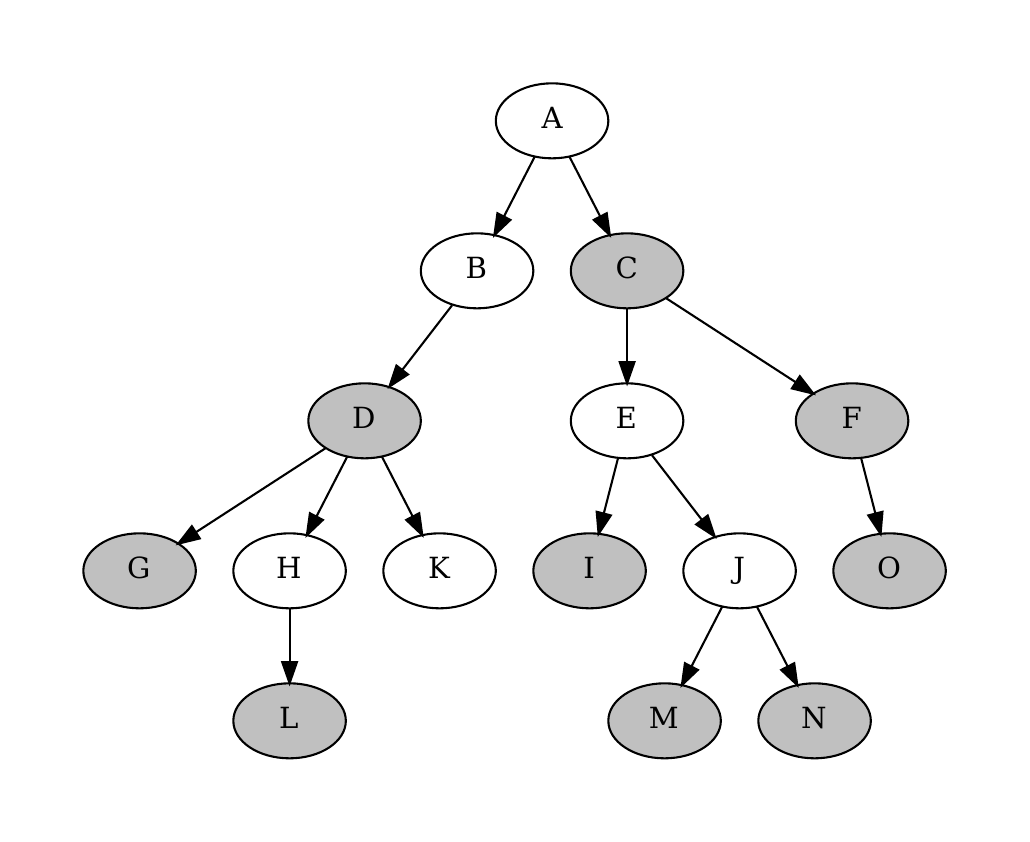}
        \label{fig:synthetic4}
    }
    \hfill
    \subfigure[Structure of synthetic SPBNs 5 and 7.]{
        \includegraphics[width=0.18\linewidth]{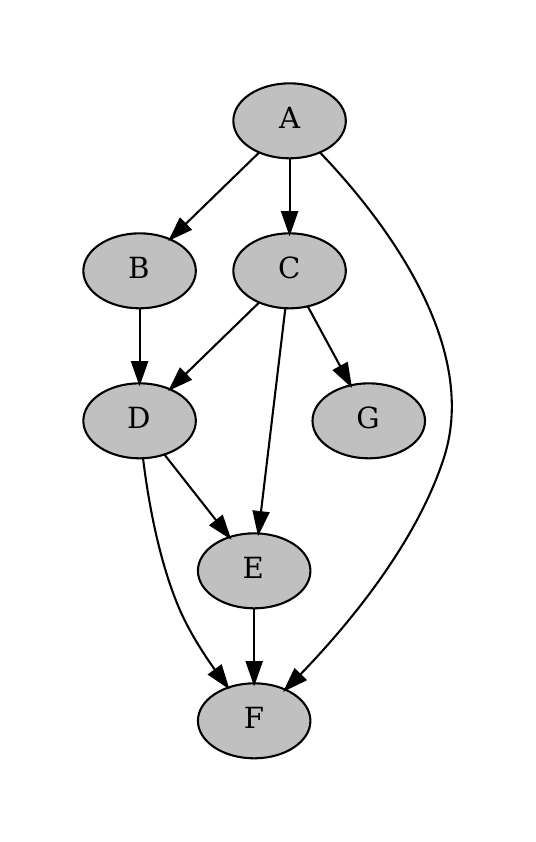}
        \label{fig:synthetic57}
    }    
    \hfill
    \subfigure[Structure of synthetic SPBNs 6 and 8.]{
        \includegraphics[width=0.24\linewidth]{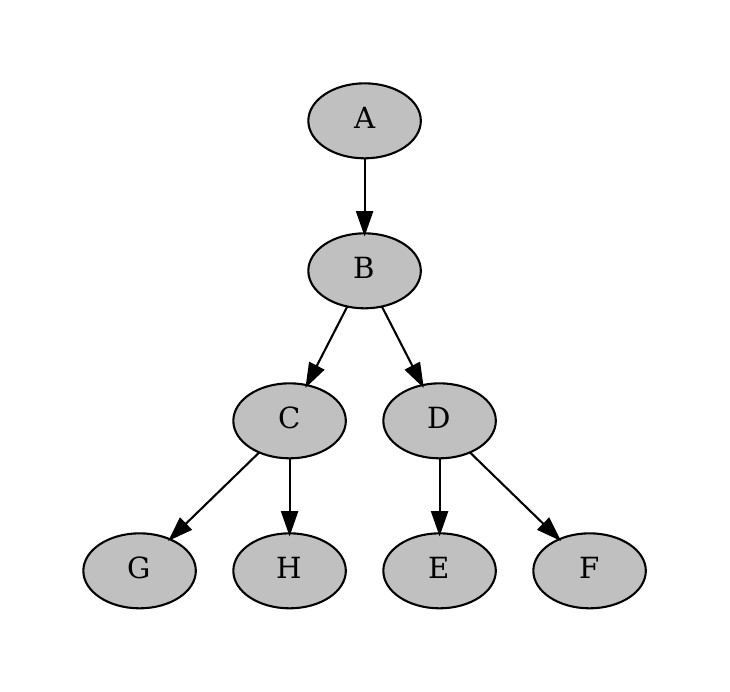}
        \label{fig:synthetic68}
    }
    \caption{Synthetic SPBNs structures.}
    \label{fig:synthetics}
\end{figure}

For the analysis of the B-SPBNs, we have created eight different probabilistic models from which to sample instances.
Four of these (Figures \ref{fig:synthetic1}, \ref{fig:synthetic2}, \ref{fig:synthetic3}, and \ref{fig:synthetic4}) are SPBNs, as they contain both LG and CKDE CPDs generated from Gaussians and mixtures of Gaussians. With them, we aim to simulate real-world scenarios where both parametric and nonparametric distributions are likely to appear as a result of the data generation process in a given problem. They also allow us to evaluate how B-SPBNs respond to different structural complexities. The other four (Figures \ref{fig:synthetic57} and \ref{fig:synthetic68}) share the arc structure of synthetic SPBNs 1 and 3, respectively, but contain only non-normal variables, where all nodes are generated from a different variety of distributions. In synthetic SPBN 5, all nodes are exponential; in synthetic SPBN 6, all nodes are gamma; in synthetic SPBN 7, all nodes are beta; and in synthetic SPBN 8, all nodes are Laplace. This will allow a thorough assessment of B-SPBN's robustness across different and more complex distributional scenarios.
Figure \ref{fig:synthetics} illustrates the structures of the corresponding SPBNs, and Table \ref{tab:synthetic_characteristics} summarizes their main characteristics. The table includes the number of nodes, the number of arcs, the maximum number of parents per node ($|\text{Pa}(i)|$), and the distributions from which the CPDs are generated.

\begin{table}[!ht]
  \centering
  \begin{tabular}{ccccl}
    \toprule
    \textbf{Model} & \textbf{Nodes} & \textbf{Arcs} & \textbf{Max $|\text{Pa}(i)|$}  & \textbf{CPDs}\\
    \midrule
     1 & 7 & 10 & 3 & Gaussian and mixtures of Gaussians\\
     2 & 13 & 21 & 5 & Gaussian and mixtures of Gaussians\\
     3 & 8 & 7 & 1 & Gaussian and mixtures of Gaussians\\
     4 & 15 & 14 & 1 & Gaussian and mixtures of Gaussians\\
     5 & 7 & 10 & 3 & Exponential\\
     6 & 8 & 7 & 1 & Gamma\\
     7 & 7 & 10 & 3 & Beta\\
     8 & 8 & 7 & 1 & Laplace\\
     
     \bottomrule
    \end{tabular}
    \caption{Characteristics of the synthetic SPBNs.}\label{tab:synthetic_characteristics}
\end{table}

The synthetic experiments have been performed from two perspectives to evaluate how B-SPBNs respond to changes in the grid size and the number of instances: 
\begin{itemize}
    \item For a fixed grid size of $M=100$ with an increasing number of training instances.
    \item For a fixed number of $N_{\text{train}} = 16384$ training instances with an increasing grid size.
\end{itemize}

For each setting, we also have two types of evaluation, where different  distance metrics will be computed. Since we are in a Bayesian network context where the ground-truth is known (Figure \ref{fig:synthetics}), we should consider the error in the structure as well as in the density estimation during the learning process.
For the network structure, these metrics include the Hamming distance (HMD), the structural Hamming distance (SHD) \citep{Tsamardinos2006}, and the node-type Hamming distance (THMD) \citep{SPBN}. The HMD measures the number of arc additions and deletions required to transform one DAG into another, ignoring the directions of the arcs. In contrast, the SHD accounts for the directional differences by also counting the number of arc flips. Similarly, the THMD captures node type differences, distinguishing between parametric and nonparametric nodes. 
For the evaluation of the JPD, we will sample a separate test dataset of size \( N_{\text{test}} = 2048 \) and compute the log-likelihood (Equation \ref{eq:logl}) of each instance. Then, the estimation error ($\hat{x}^j$ for $x^j$) will be measured using the root mean square error (RMSE) and the relative mean absolute error (RMAE) expressed in percentage:
\begin{equation}
    \text{RMSE} = \sqrt{\frac{1}{N_{\text{test}}} \sum_{j=1}^{N_{\text{test}}} (\hat{x}^j-x^j)^2} \ , \hspace{0.5cm}
    \text{RMAE}(\%) = \frac{1}{N_{\text{test}}} \sum_{j=1}^{N_{\text{test}}}  \left|\frac{\hat{x}^j-x^j}{x^j}\right  |\cdot 100
\end{equation}

To accurately perform this evaluation, the log-likelihood RMSE and RMAE(\%) of the test datasets will be computed based on the structure of the true DAG, i.e., using the structure of the corresponding synthetic SPBN (Figure \ref{fig:synthetics}) to compute the log-likelihood of each test dataset. This ensures that the evaluation remains unbiased by the arcs encountered during structure learning. 
Additionally, we will return the execution times of the B-SPBNs and the SPBNs during the running of the HC algorithm and the computation of the log-likelihood. These execution times are reported as ratios to measure the speedup of our proposal, calculated as the SPBN time divided by the B-SPBN time. Therefore, time ratios over 1 mean faster executions than the SPBN, and time ratios below 1 mean slower executions than the SPBN. For the structure learning of the network, these ratios are referred to as the HC ratio, and for the computation of the log-likelihood, they are referred to as the test ratio.

Finally, to evaluate how B-SPBNs constructed out of SBKDE or FKDE CPDs respond to the above metrics under different configurations, we will compare our proposal with the baseline using different binning rules. Simple and linear binning are the two most common fixed-width binning rules. Other high-order binning rules \citep{minnote98_binning, hall1996accuracy} extend the basis for linear binning to more adjacent bins. As we explained in Section \ref{sec:computational_cost}, linear binning increases the number of weight points, potentially leading to higher computational times in multivariate scenarios. To avoid excessive execution times, we will focus on the simple and linear binning rules. Then, the conclusions derived for the linear binning can be applied to other fixed-width high-order rules. Thus, the algorithms involved in the evaluation are:
\begin{itemize}
    \item \textbf{SPBN}. A SPBN with LG and CKDE CPDs.
    \item \textbf{B-SPBN-Simple}. A B-SPBN with LG and SBKDE CPDs using simple binning.
    \item \textbf{B-SPBN-Linear}. A B-SPBN with LG and SBKDE CPDs using linear binning.
    \item \textbf{B-SPBN-FKDE-Simple}. A B-SPBN with LG and FKDE CPDs using simple binning.
    \item \textbf{B-SPBN-FKDE-Linear}. A B-SPBN with LG and FKDE CPDs using linear binning.
\end{itemize}

\subsubsection{Log-likelihood error}
\label{sec:log-error}
\begin{figure}[H]
    \centering

    \subfigure[RMSE.]{
        \begin{minipage}{0.93\linewidth}
            \centering
            \includegraphics[width=\linewidth]{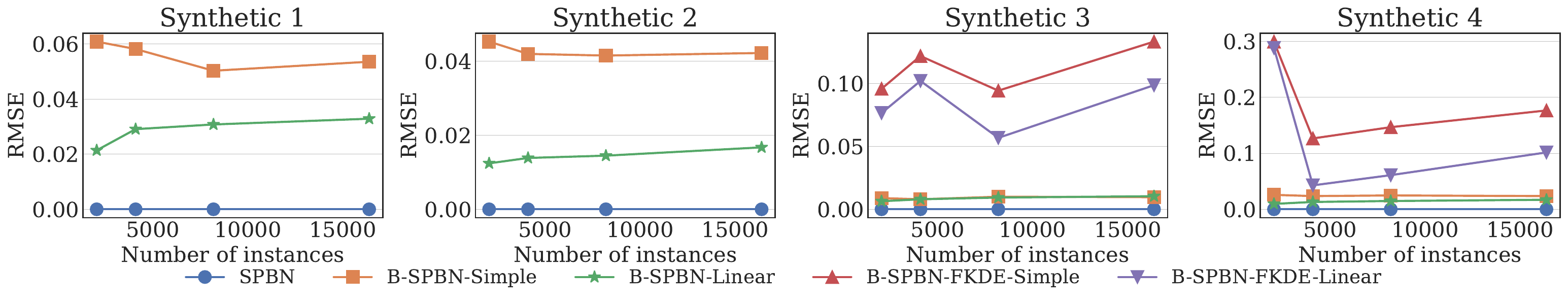}\par
            \vspace{0.25em}
            \includegraphics[width=\linewidth]{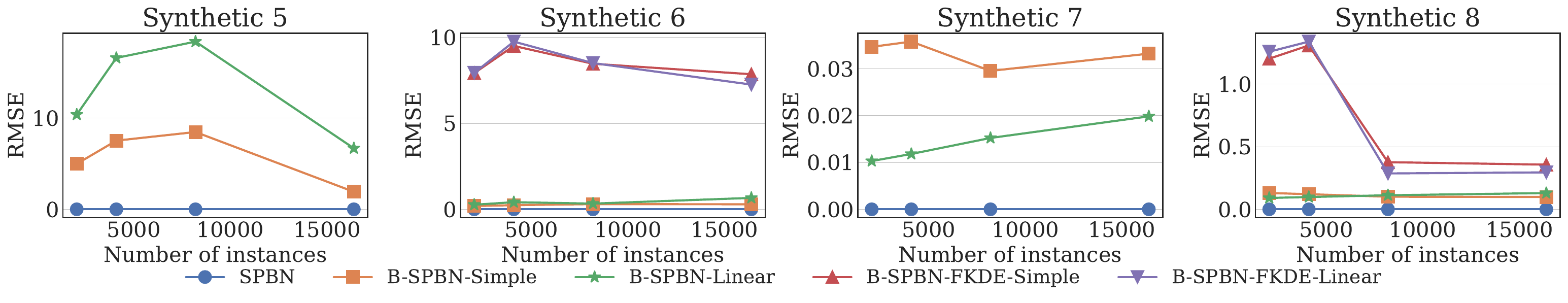}
            \label{fig:m100_rmse}
        \end{minipage}
    }
    
    \subfigure[RMAE (\%).]{
        \begin{minipage}{0.93\linewidth}
            \centering
            \includegraphics[width=\linewidth]{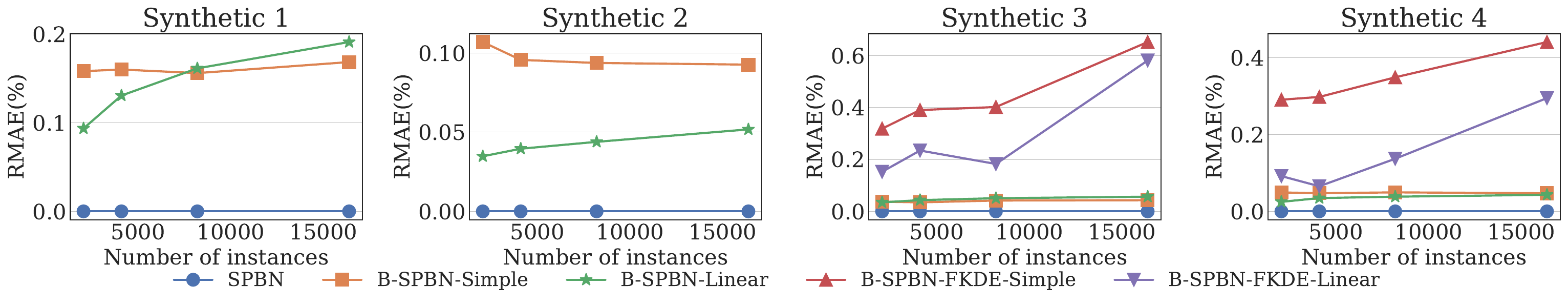}\par
            \vspace{0.25em}
            \includegraphics[width=\linewidth]{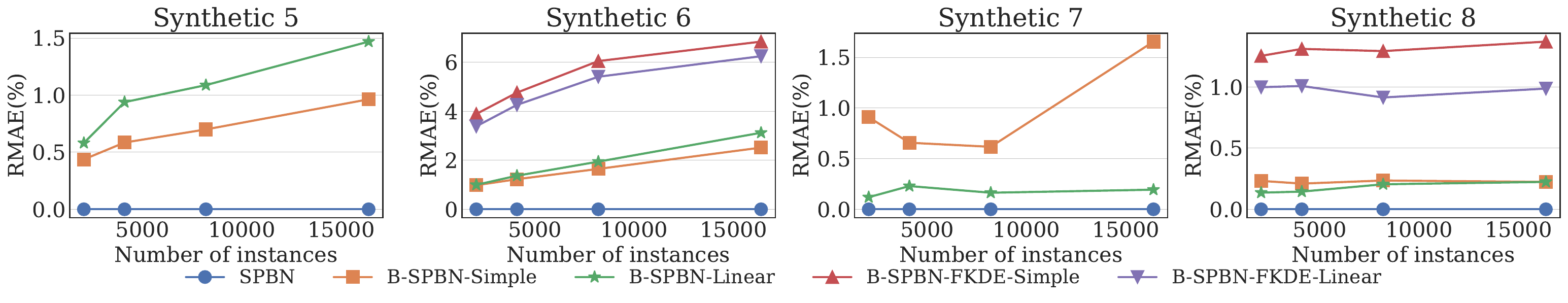}
            \label{fig:m100_rmae}
        \end{minipage}
    }
    
    \subfigure[Test ratio.]{
        \begin{minipage}{0.93\linewidth}
            \centering
            \includegraphics[width=\linewidth]{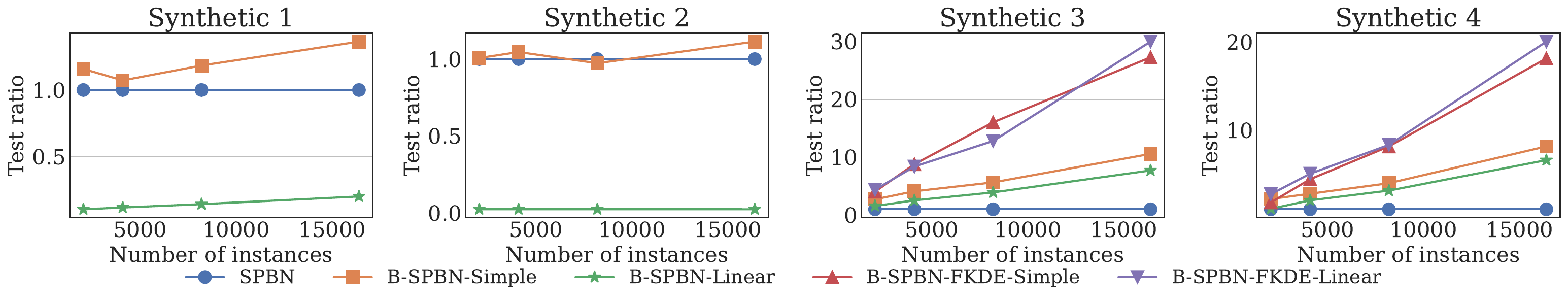}\par
            \vspace{0.25em}
            \includegraphics[width=\linewidth]{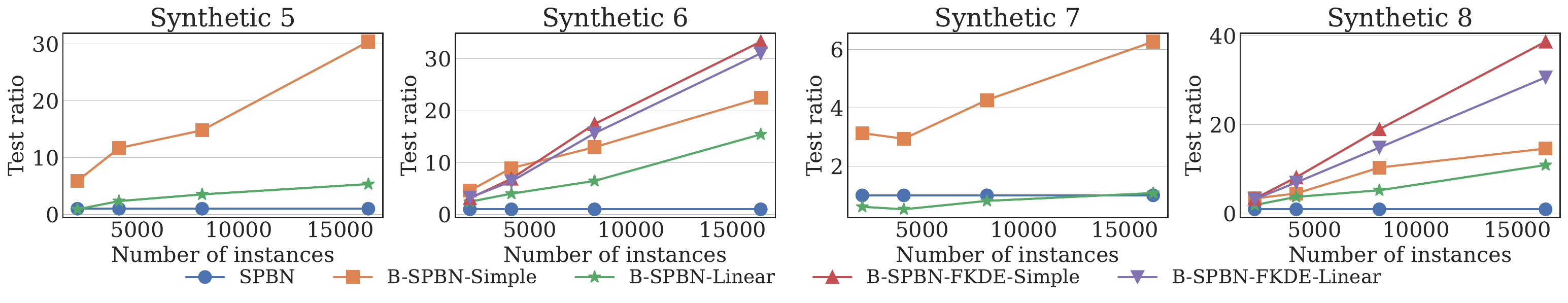}
            \label{fig:m100_test}
        \end{minipage}
    }

    \caption{Log-likelihood error results for a grid of $M=100$.}
    \label{fig:samedag100}
\end{figure}

\begin{figure}[H]
    \centering

    \subfigure[RMSE.]{
        \begin{minipage}{0.93\linewidth}
            \centering
            \includegraphics[width=\linewidth]{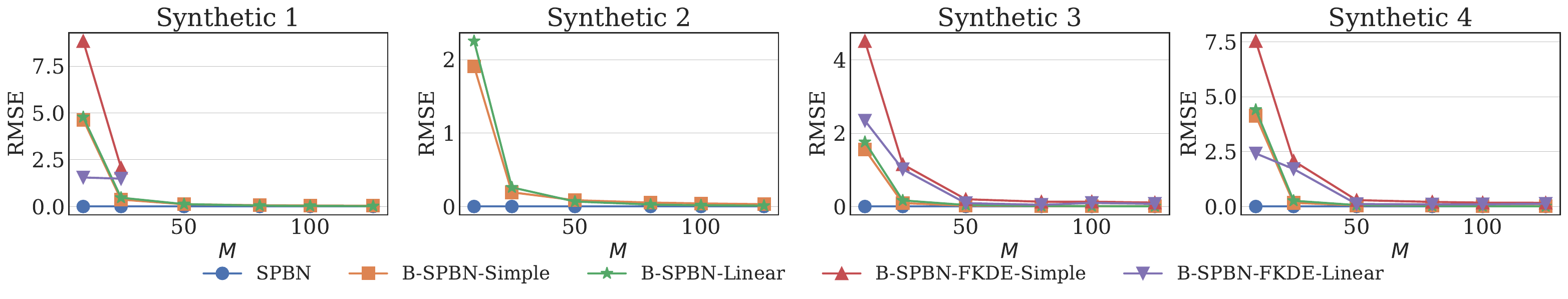}\par
            \vspace{0.25em}
            \includegraphics[width=\linewidth]{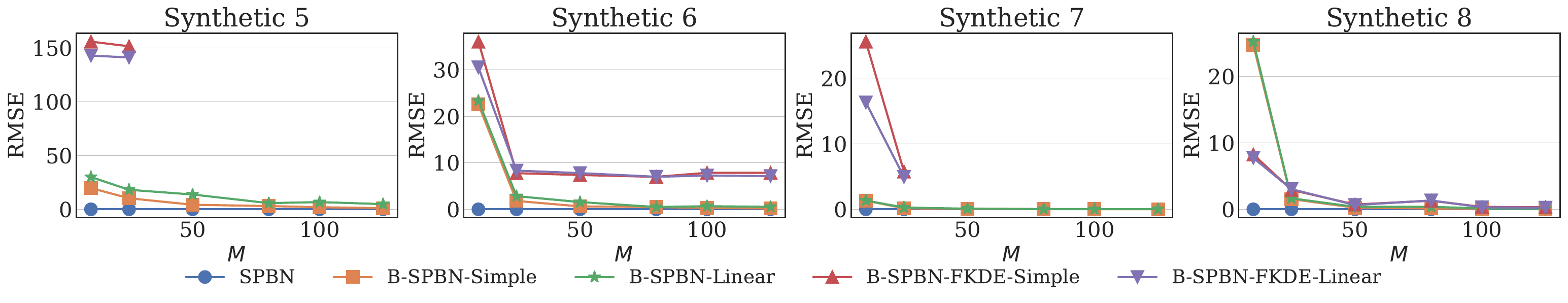}
            \label{fig:nfix_rmse}
        \end{minipage}
    }

    \subfigure[RMAE (\%).]{
        \begin{minipage}{0.93\linewidth}
            \centering
            \includegraphics[width=\linewidth]{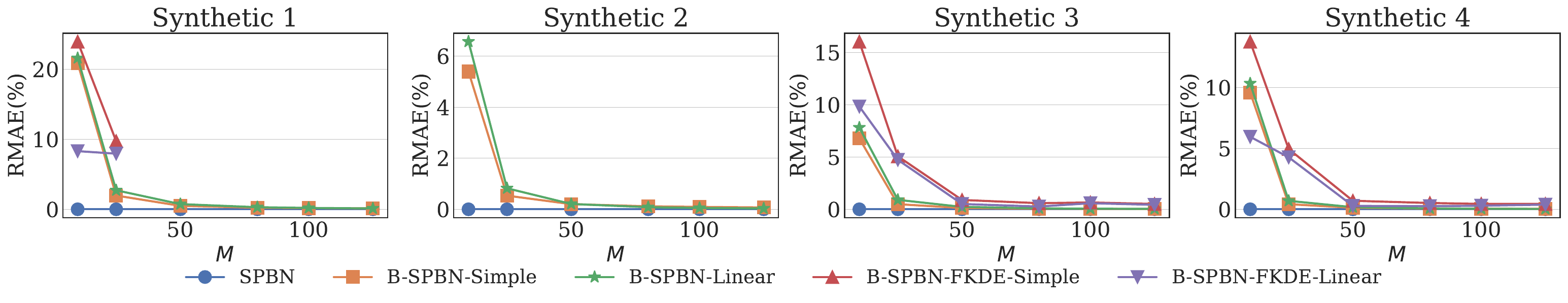}\par
            \vspace{0.25em}
            \includegraphics[width=\linewidth]{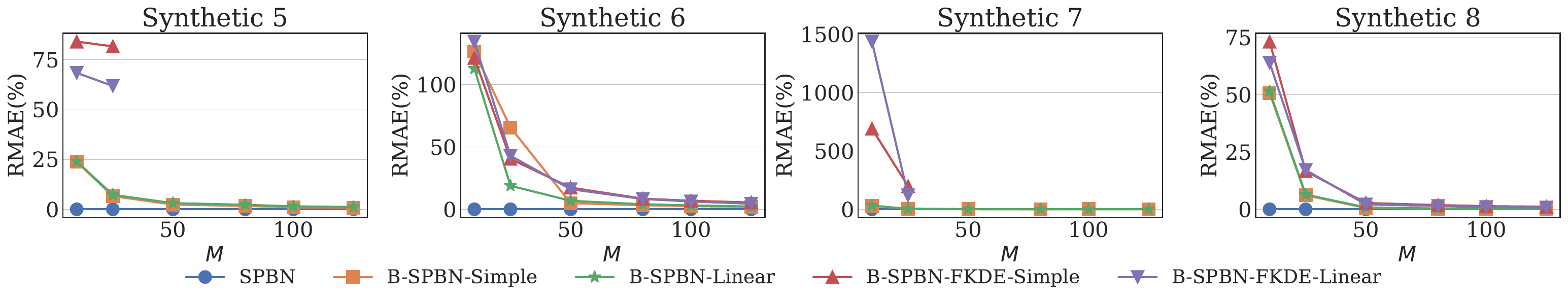}
            \label{fig:nfix_rmae}
        \end{minipage}
    }

    \subfigure[Test ratio.]{
        \begin{minipage}{0.93\linewidth}
            \centering
            \includegraphics[width=\linewidth]{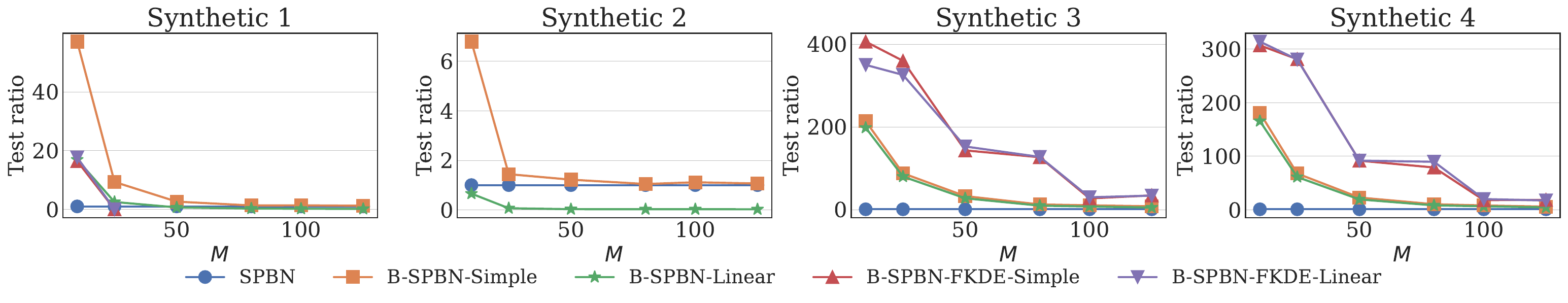}\par
            \vspace{0.25em}
            \includegraphics[width=\linewidth]{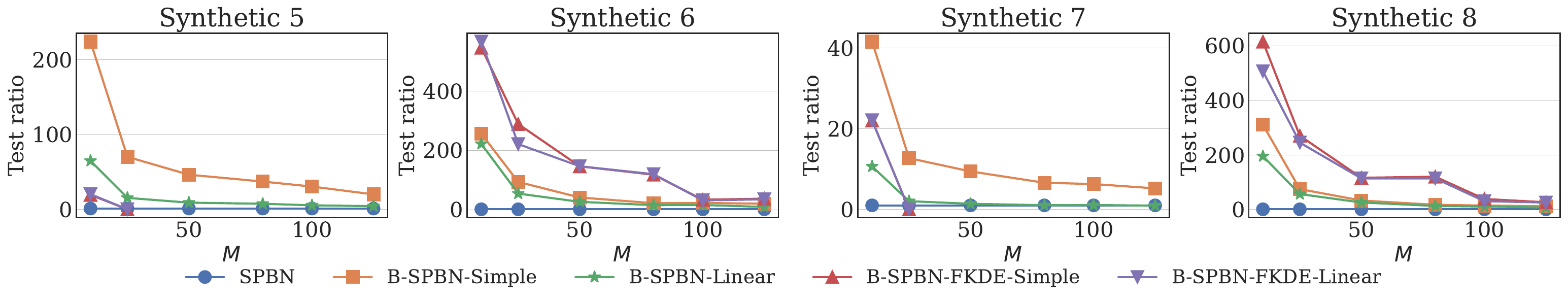}
            \label{fig:nfix_test}
        \end{minipage}
    }

    \caption{Log-likelihood error results for $N_{\text{train}} = 16384$ training instances.}
    \label{fig:samedag_Nfix}
\end{figure}

Here we show the log-likelihood error results for all synthetic SPBNs, which are illustrated in Figure \ref{fig:samedag100} (for a grid of $M=100$) and Figure \ref{fig:samedag_Nfix} (for $N_{\text{train}} = 16384$ training instances).

It can be observed that models using linear binning generally exhibited slightly lower errors than their counterparts using simple binning. Thus, B-SPBN-FKDE-Linear outperforms B-SPBN-FKDE-Simple, and B-SPBN-Linear outperforms B-SPBN-Simple in Figure \ref{fig:samedag100}. Nevertheless, the differences between B-SPBN-Simple and B-SPBN-Linear are in most cases negligible compared to B-SPBN-FKDE-Simple and B-SPBN-FKDE-Linear, especially for synthetic SPBNs 3, 4, and 8. The only exception is the RMSE of synthetic SPBN 5 (Figure \ref{fig:m100_rmse}), where, in fact, B-SPBN-Simple outperforms B-SPBN-Linear.
Also, it is worth noting that the RMSE of synthetic SPBNs 5 and 6 is much worse than the rest. The CPDs of these models follow exponential and gamma distributions. Therefore, this indicates that B-SPBNs require narrower binwidths (larger grid sizes) to accurately estimate exponential-like or highly skewed, peaked distributions.

Additionally, the times for synthetic SPBNs 1 and 2 in B-SPBN-Linear are much higher than in B-SPBN-Simple and SPBN models. Specifically, the B-SPBN-Simple achieved improvements of 10\% to 40\% compared to the baseline in synthetic SPBNs 1 and 2, and ratios with values up 30 and 6 for synthetic SPBNs 5 and 7, respectively (Figure \ref{fig:m100_test}). This behavior can be explained through the complexity analysis presented in Section \ref{sec:computational_cost}. Linear binning distributes weights across the surrounding grid points. Consequently, a node with two parents requires computations over 8 grid points per instance, while a node with five parents requires computations over 64 grid points.
For simpler structures with fewer parent nodes, such as synthetic SPBNs 3, 4, 6, and 8, Figure \ref{fig:m100_test} shows log-likelihood estimations 10 to 20 times faster than the SPBN for networks with SBKDE CPDs, and 20 to 40 times faster for those with FKDE CPDs. 

To evaluate the grid size effects on the log-likelihood error for a fixed number of instances, we used $M=10, 25, 50, 80, 100 \text{ and } 125$ (Figure \ref{fig:samedag_Nfix}). For synthetic SPBNs 1, 5, and 7, we can see that FKDE was able to handle nodes with 3 parents until $M=25$, although the ratios are already below 1 at that point. However, except for synthetic SPBN 1 with $M=10$, where B-SPBN-FKDE-Linear achieved the best results (Figure \ref{fig:nfix_rmae}), B-SPBNs with SBKDE CPDs performed significantly better than FKDE CPDs in every case. For synthetic SPBNs 3, 4, and 8, SBKDE CPDs also achieve lower errors than FKDE CPDs, but the difference is not as significant as it is for synthetic SPBNs 1, 5, 6, and 7. 
Also, we can see that the RMSE and RMAE (\%) results for synthetic SPBNs 5, 6, 7, and 8 with $M=10$ and $M=25$ are significantly worse than for synthetic SPBNs 1, 2, 3, and 4. In Figure \ref{fig:samedag100}, we only see this difference in synthetic SPBNs 5 and 6, which means that beta and Laplace distributions are, to a smaller degree, also sensitive to small grids.
In any case, all the methods in all the synthetic SPBNs show a clear decrease of the log-likelihood RMSE and RMAE (\%) after $M=50$, where B-SPBN-Simple and B-SPBN-Linear converge to values below 0.1 and 0.3\%, respectively. As mentioned, for B-SPBN-FKDE-Simple and B-SPBN-FKDE-Linear, these errors are always higher. The reason for that is likely attributed to the data binning demands, as SBKDE CPDs only have to bin the training instances, while FKDE CPDs also have to bin the test instances.

For the FKDE CPD models with lower log-likelihood errors (around $M=80$ and higher), the test ratio reaches values around 150 for synthetic SPBN 3, 6, and 8 and around 90 for synthetic SPBN 4 (Figure \ref{fig:nfix_test}). A noteworthy aspect here is the drastic change in the trends of FKDE CPD models when the grid size increases from \( M = 25 \) to \( M = 50 \) and from \( M = 80 \) to \( M = 100 \) in Figure \ref{fig:nfix_test}. This behavior is most likely influenced by the size of \( P \), which remains constant from 50 to 80 and from 100 to 125. According to Equation (\ref{eq:size_P}), \( P = 256 \) for \( M = 50 \) and \( M = 80 \), whereas for \( M = 100 \) and \( M = 125 \), it increases to \( P = 512 \). As a result, the test ratios drop by 100 orders of magnitude, and the RMSE and \text{RMAE}(\%) slightly increase, respectively, for synthetic SPBNs 3 (Figure \ref{fig:nfix_rmae}) and 8 (Figure \ref{fig:nfix_rmse}). Possibly, this is due to the greater presence of zeros in the padding.

To summarize the analysis, we can conclude that, if the structure is known, B-SPBN-Simple performs better than B-SPBN-Linear, considering the small error difference and the higher cost of B-SPBN-Linear. On the contrary, B-SPBN-FKDE-Linear returns significantly lower errors than B-SPBN-FKDE-Simple at practically the same ratios. Between B-SPBN-Simple and SPBN-FKDE-Linear, the choice depends on the complexity of the problem, i.e., the number of parent nodes, and the error that we are willing to admit. In both cases, grid sizes higher than $M=50$ return reasonably good results for a wide variety of distributions.

\subsubsection{Structure learning error}
\begin{figure}[!htb]
    \centering

    \subfigure[$M=100$.]{
        \begin{minipage}{.95\linewidth}
            \centering
            \includegraphics[width=\linewidth]{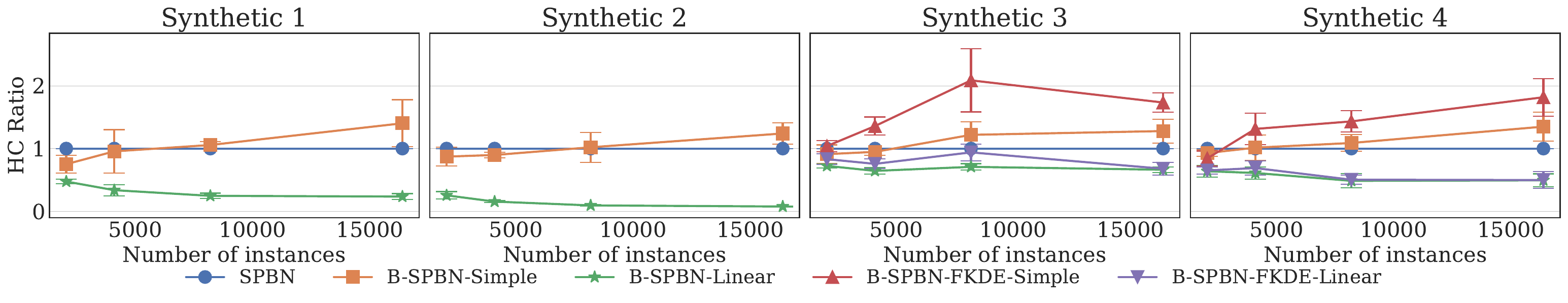}\par
            \vspace{0.25em}
            \includegraphics[width=\linewidth]{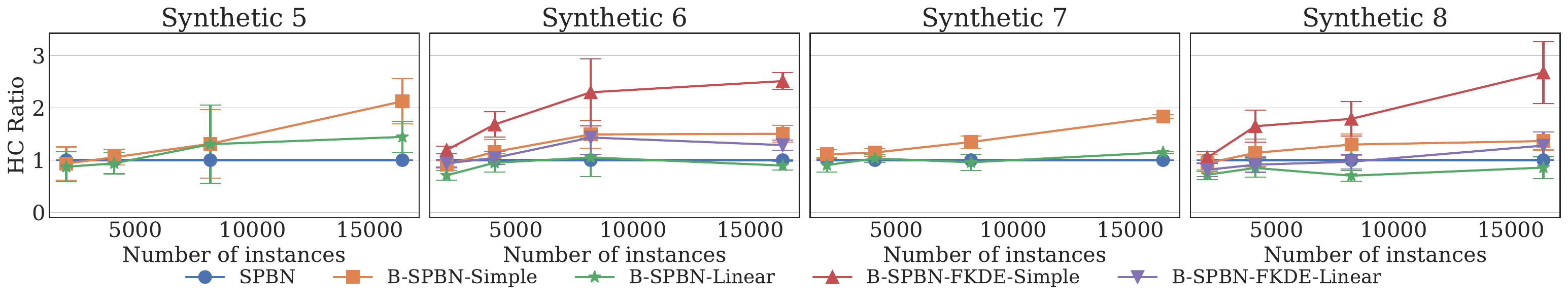}
            \label{fig:m100_hcratio}
        \end{minipage}
    }

        \subfigure[$N_{\text{train}} = 16384$.]{
        \begin{minipage}{.95\linewidth}
            \centering
            \includegraphics[width=\linewidth]{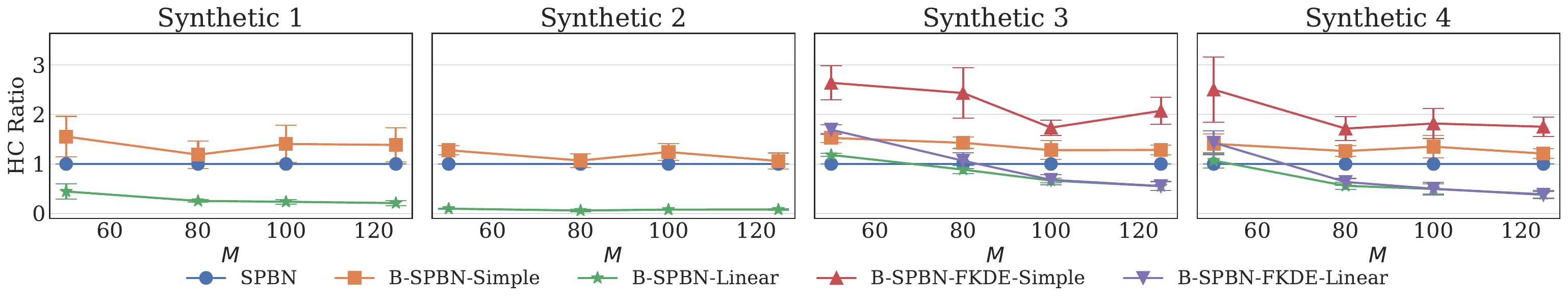}\par
            \vspace{0.25em}
            \includegraphics[width=\linewidth]{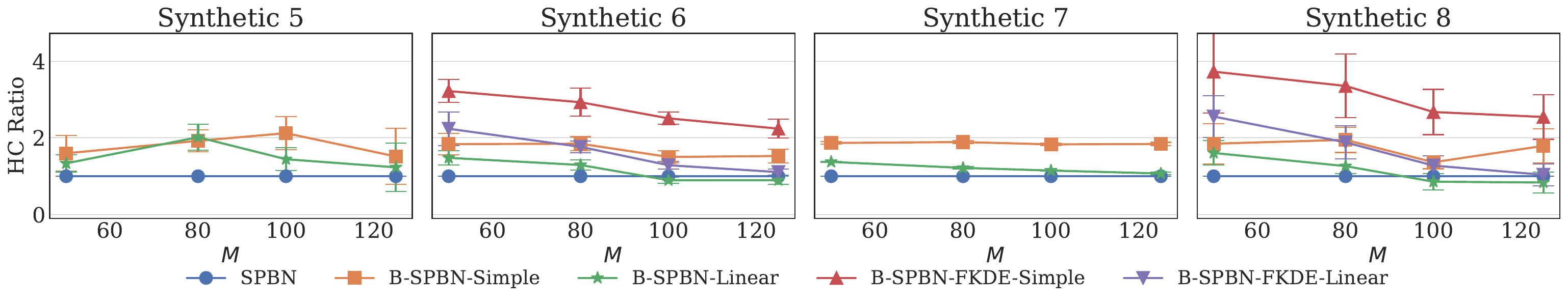}
            \label{fig:Nfix_hcratio}
        \end{minipage}
    }

    \caption{Structural learning results. HC Ratio.}
    \label{fig:hcratio_results}
\end{figure}

\begin{figure}[!ht]
    \centering

    \subfigure[HMD.]{
        \begin{minipage}{0.91\linewidth}
            \centering
            \includegraphics[width=\linewidth]{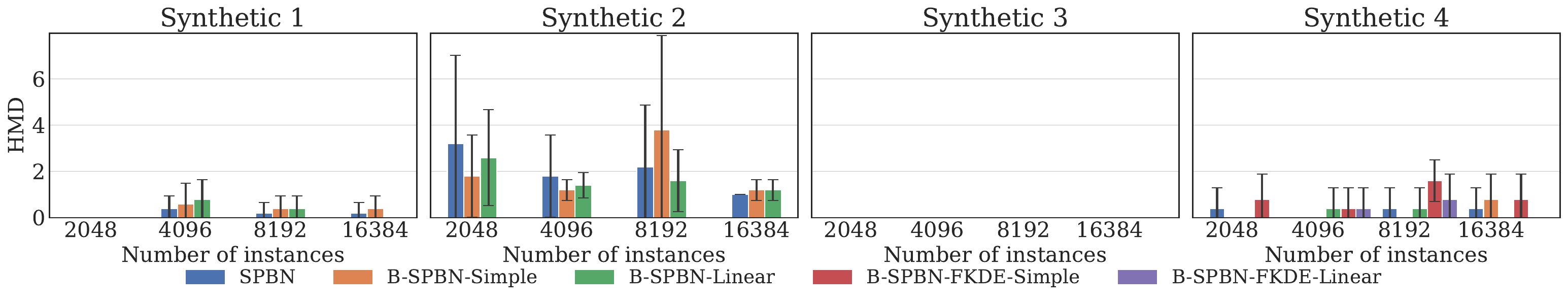}\par
            \vspace{0.25em}
            \includegraphics[width=\linewidth]{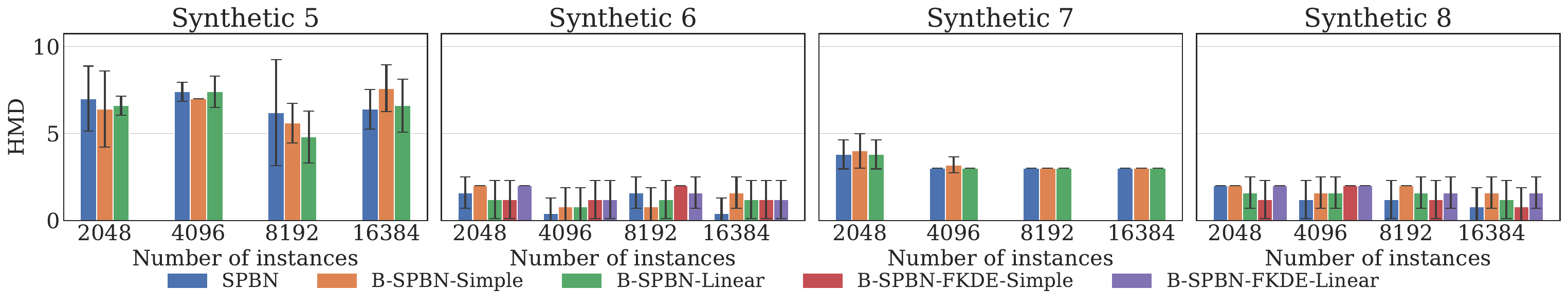}
            \label{fig:m100_hmd}
        \end{minipage}
    }\par\vspace{1em}

    \subfigure[SHD.]{
        \begin{minipage}{0.91\linewidth}
            \centering
            \includegraphics[width=\linewidth]{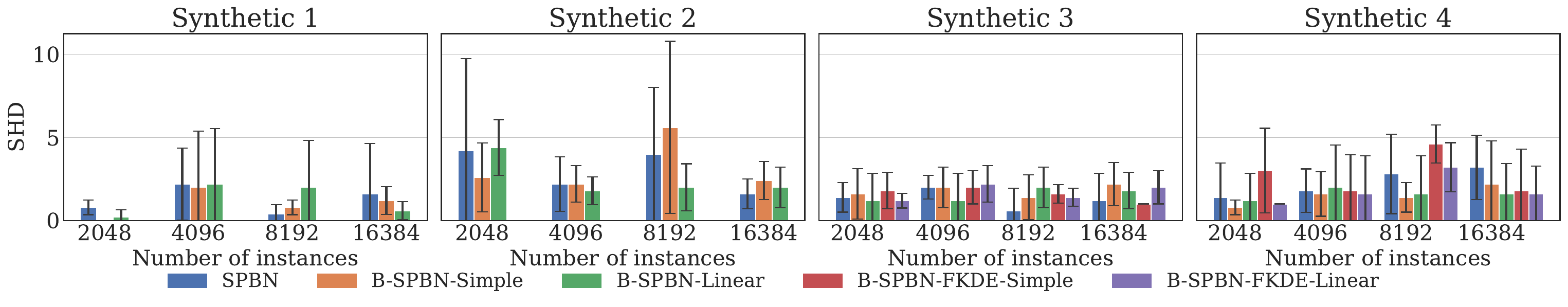}\par
            \vspace{0.25em}
            \includegraphics[width=\linewidth]{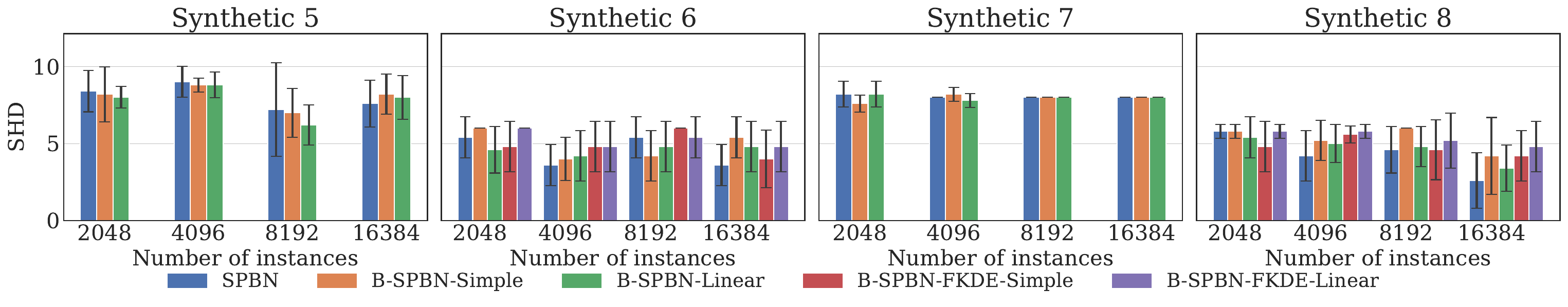}
            \label{fig:m100_shd}
        \end{minipage}
    }

    \subfigure[THMD.]{
        \begin{minipage}{0.91\linewidth}
            \centering
            \includegraphics[width=\linewidth]{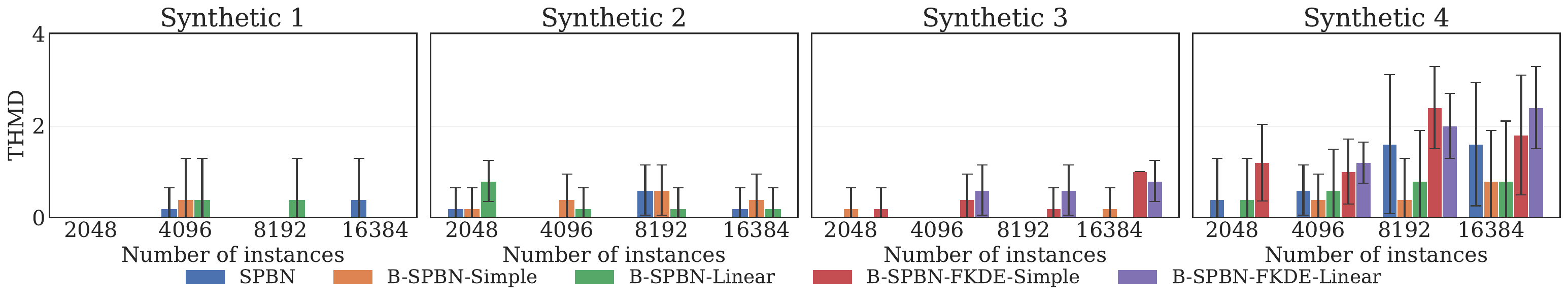}\par
            \vspace{0.25em}
            \includegraphics[width=\linewidth]{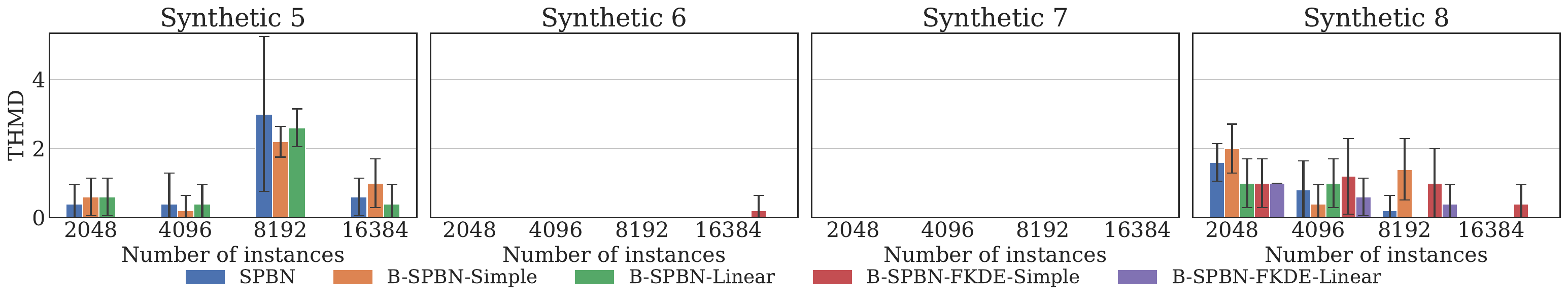}
            \label{fig:m100_thmd}
        \end{minipage}
    }

    \caption{Structural learning results for a grid of $M=100$.}
    \label{fig:estimated100}
\end{figure}

\begin{figure}[!htb]
    \centering

    \subfigure[HMD.]{
        \begin{minipage}{0.91\linewidth}
            \centering
            \includegraphics[width=\linewidth]{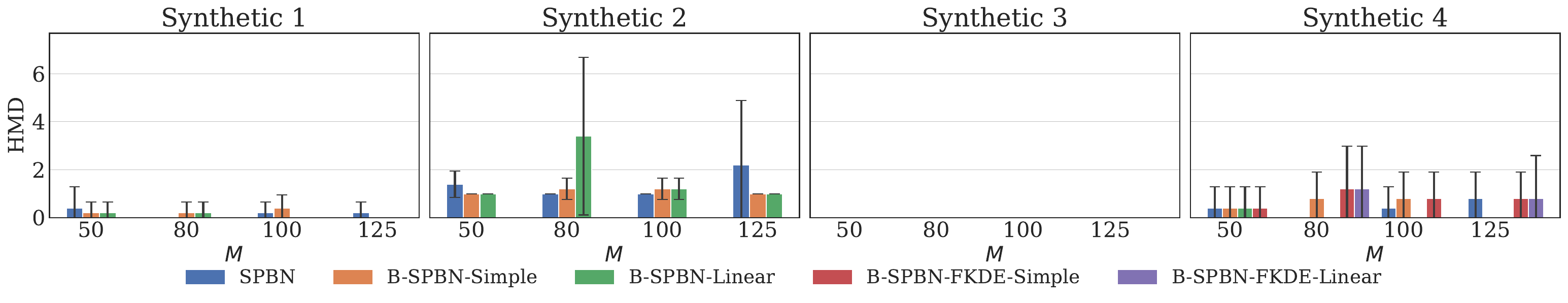}\par
            \vspace{0.5em}
            \includegraphics[width=\linewidth]{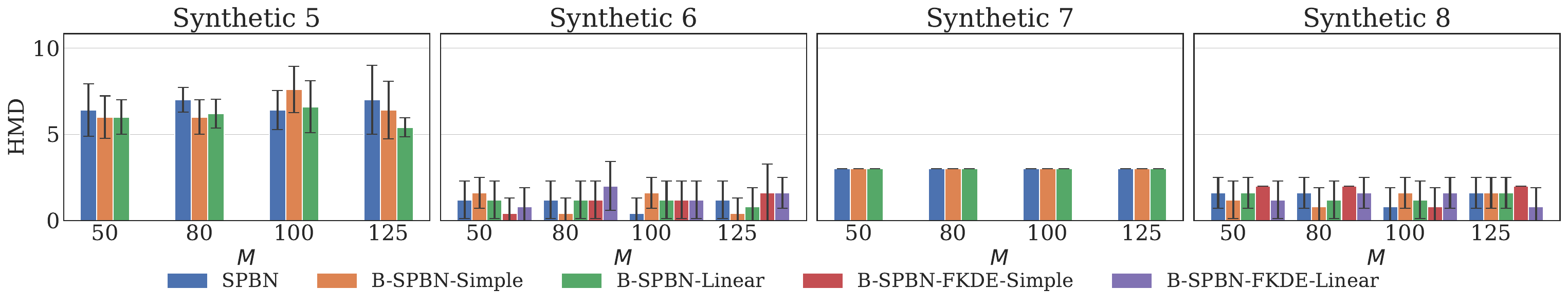}
            \label{fig:hmd_Nfix}
        \end{minipage}
    }

    \subfigure[SHD.]{
        \begin{minipage}{0.91\linewidth}
            \centering
            \includegraphics[width=\linewidth]{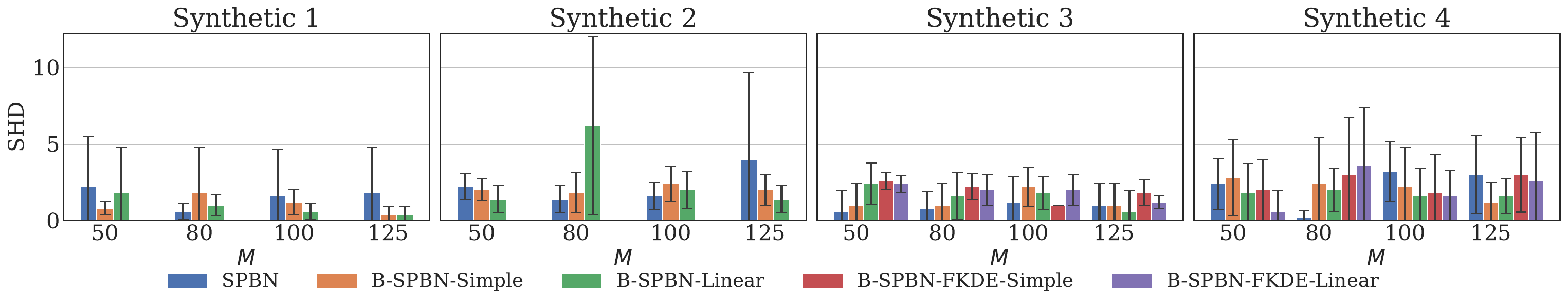}\par
            \vspace{0.25em}
            \includegraphics[width=\linewidth]{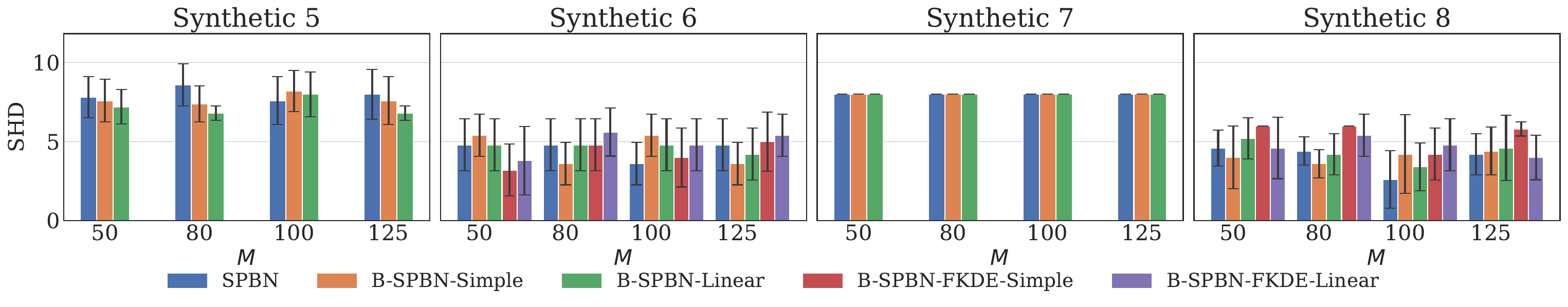}
            \label{fig:shd_Nfix}
        \end{minipage}
    }

    \subfigure[THMD.]{
        \begin{minipage}{0.91\linewidth}
            \centering
            \includegraphics[width=\linewidth]{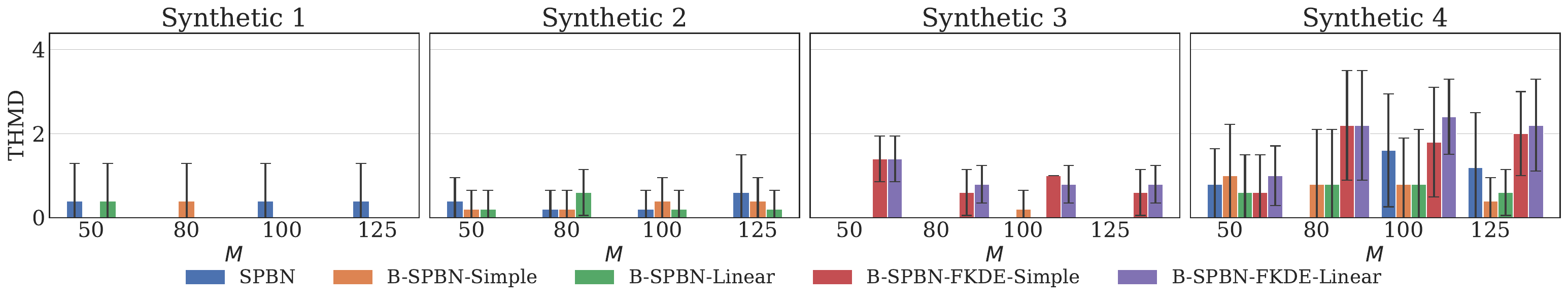}\par
            \vspace{0.25em}
            \includegraphics[width=\linewidth]{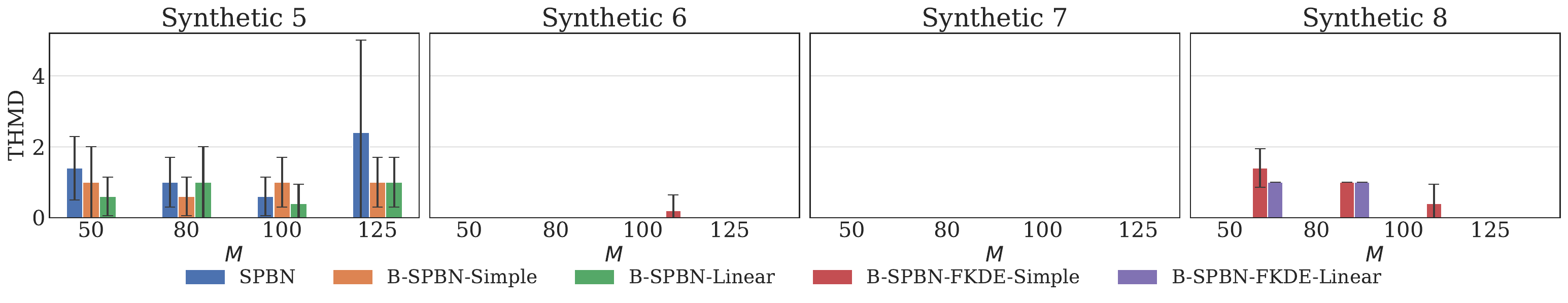}
            \label{fig:thmd_Nfix}
        \end{minipage}
    }

    \caption{Structural learning results for $N_{\text{train}} = 16384$ training instances.}
    \label{fig:estimated_Nfix}
\end{figure}

The structural learning results are presented, for the HC ratios in Figure \ref{fig:hcratio_results}, and for the different Hamming distances (HMD, SHD, THMD) as error bars in Figure \ref{fig:estimated100} (for a grid of $M=100$) and Figure \ref{fig:estimated_Nfix} (for $N_{\text{train}} = 16384$ training instances). Given the previous results, we computed Figure \ref{fig:estimated_Nfix} for grid sizes starting at $M=50$. Therefore, B-SPBN-FKDE-Simple and B-SPBN-FKDE-Linear are restricted to one parent and do not appear in the results of synthetic SPBNs 1, 2, 5, and 7.

For the structure learning process, the time ratios are lower since every change performed by HC during the network estimation requires data binning. At a lower number of instances, as well as for the linear binning rule, this extra time is more significant, see Figure \ref{fig:m100_hcratio}. In fact, B-SPBN-FKDE-Linear, which previously performed at similar ratios as B-SPBN-FKDE-Simple, here is slower than the baseline in almost every case and runs at similar speeds as B-SPBN-Linear. In contrast, B-SPBN-Simple is at least 20\% to 40\%  faster for synthetic SPBNs 1, 2, 3, and 4, and 50\% to 100\% for synthetic SPBNs 5, 6, 7, and 8. Here, B-SPBN-FKDE-Simple is nearly twice as fast as the baseline for synthetic SPBNs 3 and 4, and achieves ratios of 2.5 and 2.8 for synthetic SPBNs 6 and 8. For smaller grid sizes, see Figure \ref{fig:Nfix_hcratio}, the ratios of all the methods increase as expected.

\begin{figure}[!h]
    \centering
    \subfigure[SHD.]{
        \includegraphics[width=0.9\linewidth]{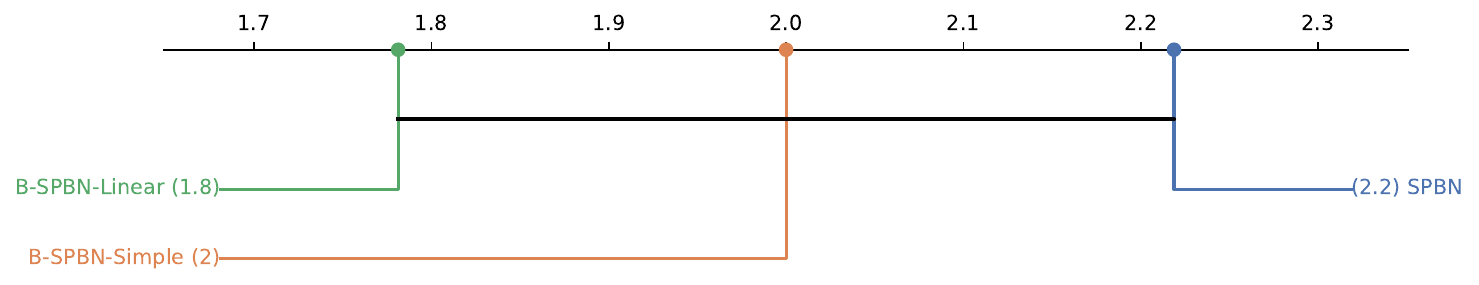}
        \label{fig:shd_bergmann_m100_s12}
    }
    \subfigure[THMD.]{
        \includegraphics[width=0.9\linewidth]{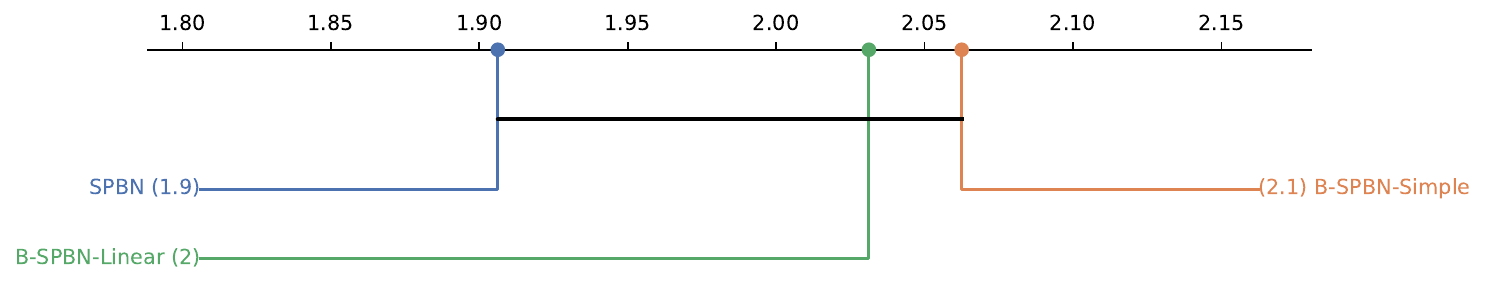}
        \label{fig:thmd_bergmann_m100_s12}
    }    
    \caption{Critical difference diagram for the SHD and THMD of synthetic datasets 1, 2, 5, and 7.}
    \label{fig:bergmann_m100_s12}
\end{figure}

\begin{figure}[!h]
    \centering
    \subfigure[SHD.]{
        \includegraphics[width=0.95\linewidth]{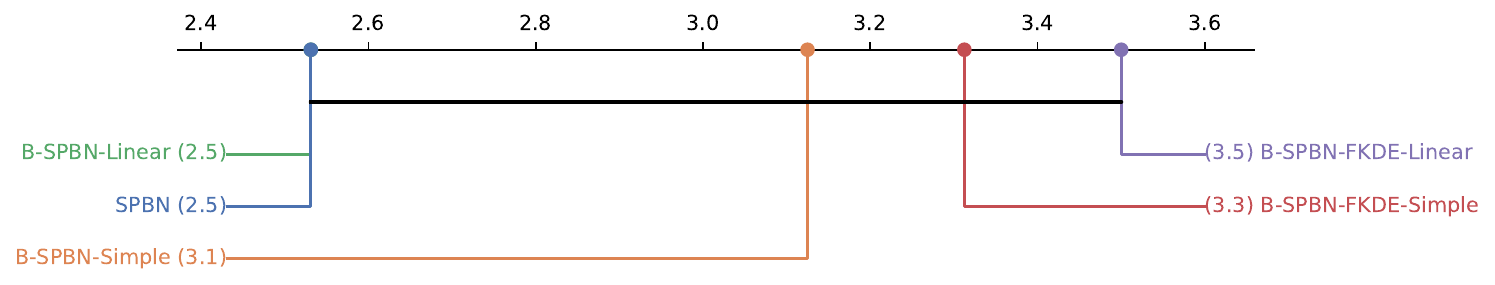}
        \label{fig:shd_bergmann_m100_s34}
    }
    \subfigure[THMD.]{
        \includegraphics[width=0.95\linewidth]{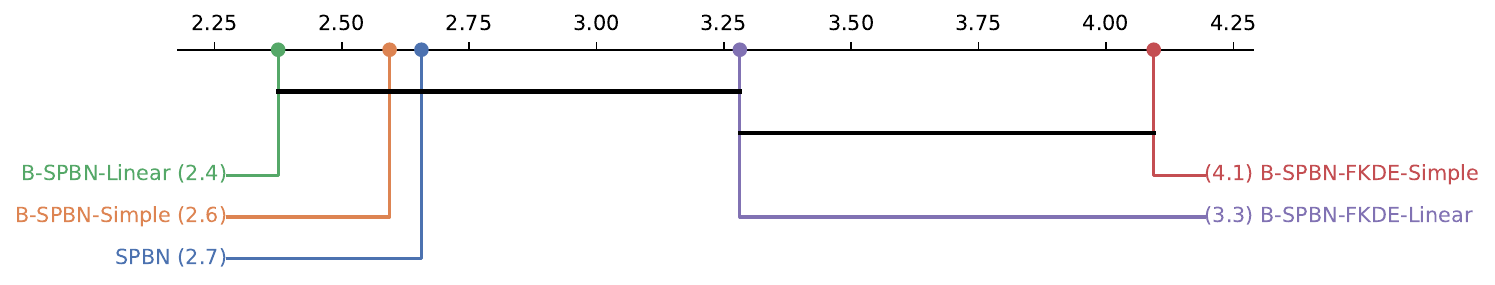}
        \label{fig:thmd_bergmann_m100_s34}
    }    
    \caption{Critical difference diagram for the SHD and THMD of synthetic datasets 3, 4, 6, and 8.}
    \label{fig:bergmann_m100_s34}
\end{figure}

To analyze the errors, we performed a Friedman test with a significance level of $\alpha = 0.05$, followed by a Bergmann-Hommel \textit{post-hoc} analysis \citep{friedman_with_berg-hommel} to identify pairwise significant structural differences between the new B-SPBNs and the standard SPBNs. The results, shown in Figure \ref{fig:bergmann_m100_s12} and Figure \ref{fig:bergmann_m100_s34}, are presented in a critical difference diagram \citep{critical_diff}, where the horizontal lines connect models without significant differences. The models are sorted from left to right according to their mean rank (the number in parentheses). The better the model, the lower the number. 
Given that the SHD incorporates arc flips, while the HMD does not, we used the SHM and THMD metrics for the analysis. Since the B-SPBN-FKDE-Simple and B-SPBN-FKDE-Linear models are restricted to one parent, the evaluation has been done separately for synthetic SPBNs 1, 2, 5, and 7 and synthetic SPBNs 3, 4, 6, and 8.
Thus, the results indicate no statistically significant differences between the networks concerning the SHD. These results align with those shown in Figure \ref{fig:m100_shd} and Figure \ref{fig:shd_Nfix}, as the bar heights are approximately the same. In contrast, the critical difference diagram for the THMD shows that B-SPBNs with FKDE CPDs are significantly worse than the other models in accurately determining the node type, see Figure \ref{fig:thmd_bergmann_m100_s34}. This is also reflected in Figure \ref{fig:m100_thmd}, and especially in Figure \ref{fig:thmd_Nfix}.

As in the previous section, the simple binning rule is more computationally efficient, since there are no significant differences between the structures of B-SPBN-Simple, B-SPBN-Linear, and the baseline. In contrast, for B-SPBN-FKDE-Linear, the data binning process increased the structure learning time, penalizing the HC ratios compared to the test ratios shown previously. In terms of structure, there are no significant differences with respect to B-SPBN-FKDE-Simple except for the node type, where B-SPBN-FKDE-Linear performed notably better according to the ranking. 

\subsection{UCI Machine Learning repository}
For the experiments using data from the UCI Machine Learning repository, we selected seven unlabeled datasets with continuous variables from different domains. We removed the timestamps, discrete columns, and null values. Table \ref{tab:real_datasets} presents the datasets along with their characteristics after the preprocessing.

\begin{table}[!ht]
  \centering
  \begin{tabular}{clrr}
    \toprule
    \textbf{Dataset} & \textbf{Name}  &\textbf{$N$} & \textbf{$n$}  \\
    \midrule
     1 & Single elder home monitoring: gas and position \citep{1_single_elder}                                      & 416153  & 9  \\
     2 & HTRU2 \citep{2_htru2}                                                                                      & 17898   & 8 \\
     3 & Individual household electric power consumption \citep{individual_household_electric_power_consumption_235} & 2049280 & 7  \\
     4 & MAGIC gamma telescope \citep{4_magic_gamma_telescope}                                                      & 19020   & 10 \\
     5 & Appliances energy prediction \citep{5_appliance_energy_predicion}                                          & 19735   & 24 \\
     6 & Gas turbine CO and NOx emission data set \citep{6_gas_turbine}                       & 36733   & 11 \\
     7 & Gas sensor array under dynamic gas mixtures \citep{7_ethylene_co}                     & 4178504   & 16 \\
     \bottomrule
    \end{tabular}
    \caption{Datasets from the UCI Machine Learning repository.}\label{tab:real_datasets}
\end{table}

For the comparison against other models, we included Gaussian Bayesian networks (GBNs) with two commonly known scores, the Bayesian information criterion (BIC) \citep{koller2009probabilistic} and the Bayesian Gaussian equivalent (BGe) \citep{bge_score}. 
As mentioned at the beginning of Section \ref{sec:experiments}, these experiments are conducted on a continuous Bayesian network basis. To highlight the differences between our proposal and other types of networks, we decided to include one of the most popular types, the GBNs. 
However, GBNs are parametric models, which makes them significantly faster to learn from data due to the Gaussianity assumption. Therefore, these were excluded from the time analysis, since the B-SPBNs aim to optimize SPBNs by reducing the computational cost of CKDE CPDs, and GBNs are assumed to be faster than any nonparametric or semiparametric model.
For computational efficiency, we also excluded B-SPBN-Linear and B-SPBN-FKDE-Linear, since according to Section \ref{sec:synthetic_datasets}, they only return slightly better log-likelihoods and do not show significant structural differences with respect to the simple binning methods.

\begin{figure}[!h]
    \centering
    \subfigure[Only one parent $(|\text{Pa}(i)|=1)$.]{
        \includegraphics[width=0.915\linewidth]{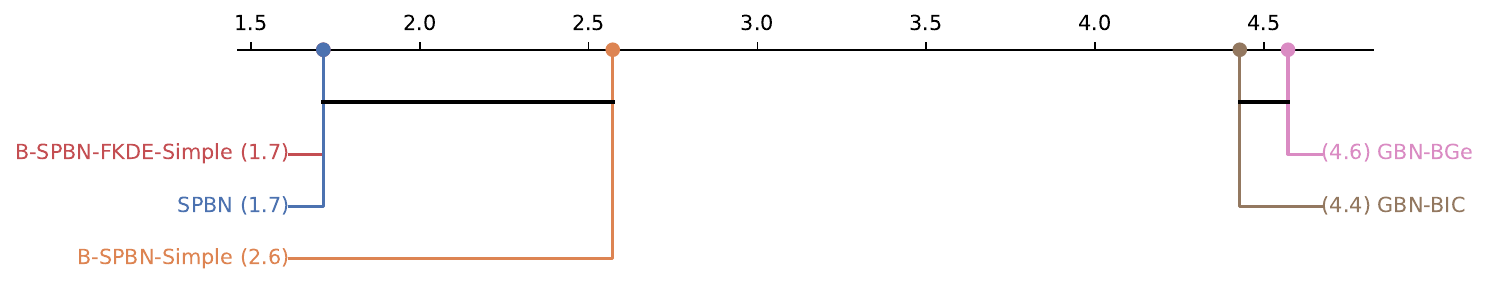}
        \label{fig:slogl_1pa}
    }
    \subfigure[More than one parent $(|\text{Pa}(i)|>1)$.]{
        \includegraphics[width=0.915\linewidth]{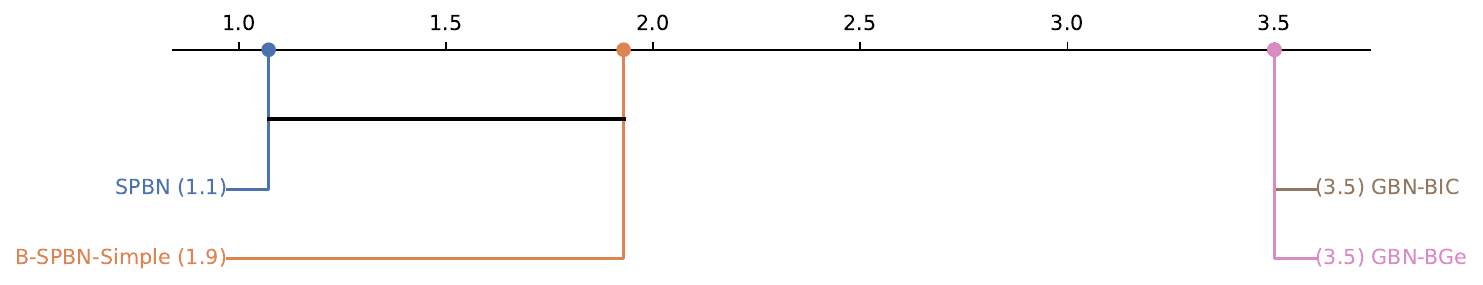}
        \label{fig:slogl_no_pa}
    }

    \caption{Critical difference diagram for the log-likelihood of the real datasets.}
    \label{fig:bergmann_real}
\end{figure}

For the evaluation of the datasets, we sampled $N_{\text{train}} = 16384$ training instances and $N_{\text{test}} = 2048$ test instances. Then, we conducted experiments for two grid sizes, \( M = 50 \) and \( M = 100 \), using a single parent node (\(|\text{Pa}(i)| = 1\)) and multiple parent nodes (\(|\text{Pa}(i)| > 1\)). 
In this case, the underlying structure of the data is unknown, so we cannot use the HMD, SHD, and THMD metrics. Therefore, we compared the models using the log-likelihood of the test datasets and a Friedman test with the Bergmann-Hommel \textit{post-hoc} analysis to identify statistically significant differences. For the time evaluation, we used the HC and test ratios as before. The Bergmann-Hommel analysis of the log-likelihood is shown in Figure \ref{fig:bergmann_real}, while Table \ref{tab:results_real} presents the average time ratios and standard deviations of the B-SPBNs.

\begin{table}[!ht]
  \centering
  \begin{tabular}{cllcccc}
\toprule                    
\textbf{Dataset} & \textbf{Model}             & \bm{$M$} & \textbf{HC Ratio}  & \textbf{HC Ratio}   & \textbf{Test Ratio }   & \textbf{Test Ratio }   \\
                 &                            &            & \textbf{$(|\text{Pa}(i)|=1)$}   &  \textbf{$(|\text{Pa}(i)|>1)$}   & \textbf{$(|\text{Pa}(i)|=1)$}       &  \textbf{$(|\text{Pa}(i)|>1)$} \\
\midrule
\multirow{4}{*}{1}  & \multirow{2}{*}{B-SPBN-Simple}      & 50         & 1.56 $\pm$ 0.17 &  1.19 $\pm$ 0.28    & 2.87e+01 $\pm$ 4.95  & 1.72 $\pm$ 0.16 \\
                    &                            & 100        & 1.48 $\pm$ 0.18 &  1.10 $\pm$ 0.17    & 9.66e+00 $\pm$ 1.19  & 1.40 $\pm$ 0.09  \\ \cmidrule{2-7}
                    & \multirow{2}{*}{B-SPBN-FKDE-Simple} & 50         & 3.87 $\pm$ 0.83 &          -          & \hspace{1.7mm}1.23e+02 $\pm$ 18.37 &           -         \\
                    &                            & 100        & 2.85 $\pm$ 0.31 &          -          & 2.37e+01 $\pm$ 1.59  &           -         \\ \hline
\multirow{4}{*}{2}  & \multirow{2}{*}{B-SPBN-Simple}      & 50         & 1.75 $\pm$ 0.03 &  1.92 $\pm$ 0.23    & 3.68e+01 $\pm$ 4.26  & \textbf{9.65 $\pm$ 1.02} \\
                    &                            & 100        & 1.62 $\pm$ 0.03 &  1.45 $\pm$ 0.05    & 1.70e+01 $\pm$ 2.07  & \textbf{3.49 $\pm$ 0.15}  \\ \cmidrule{2-7}
                    & \multirow{2}{*}{B-SPBN-FKDE-Simple} & 50         & 3.89 $\pm$ 0.19 &          -          & \hspace{1.7mm}1.23e+02 $\pm$ 24.20 &           -         \\
                    &                            & 100        & 3.11 $\pm$ 0.10 &          -          & \textbf{2.79e+01 $\pm$ 2.89}  &           -         \\ \hline
\multirow{4}{*}{3}  & \multirow{2}{*}{B-SPBN-Simple}      & 50         & 2.03 $\pm$ 0.05 & \textbf{ 2.08 $\pm$ 0.12 }   & \textbf{3.89e+01 $\pm$ 3.92}  & \hspace{-1.7mm}\textbf{14.10 $\pm$ 2.14} \\
                    &                            & 100        & \textbf{1.92 $\pm$ 0.03} &  \textbf{1.76 $\pm$ 0.18}    & \textbf{1.80e+01 $\pm$ 3.13}  & \textbf{9.83  $\pm$ 4.47} \\ \cmidrule{2-7}
                    & \multirow{2}{*}{B-SPBN-FKDE-Simple} & 50         & \textbf{4.48 $\pm$ 0.03} &          -          & \hspace{1.7mm}1.26e+02 $\pm$ 14.17 &           -         \\
                    &                            & 100        & 2.98 $\pm$ 0.04 &          -          & 3.12e+01 $\pm$ 3.03  &           -         \\ \hline
\multirow{4}{*}{4}  & \multirow{2}{*}{B-SPBN-Simple}      & 50         & 2.03 $\pm$ 0.67 &  1.39 $\pm$ 0.18    & 2.09e+01 $\pm$ 6.77  & 2.33  $\pm$ 1.34 \\
                    &                            & 100        & 1.60 $\pm$ 0.49 &  1.07 $\pm$ 0.28	  & 9.17e+00 $\pm$ 1.35  & 1.25 $\pm$ 0.24  \\ \cmidrule{2-7}
                    & \multirow{2}{*}{B-SPBN-FKDE-Simple} & 50         & \textbf{4.47 $\pm$ 0.89} &          -          & \hspace{1.7mm}1.17e+02 $\pm$ 11.84 &           -         \\
                    &                            & 100        & \textbf{3.39 $\pm$ 0.54} &          -          & 2.38e+01 $\pm$ 1.52  &           -         \\ \hline
\multirow{4}{*}{5}  & \multirow{2}{*}{B-SPBN-Simple}      & 50         & 1.90 $\pm$ 0.48 &  1.07 $\pm$ 0.16    & 3.86e+01 $\pm$ 7.67  & 1.68 $\pm$ 0.04 \\
                    &                            & 100        & 1.36 $\pm$ 0.10 &  1.09 $\pm$ 0.10    & 1.03e+01 $\pm$ 0.25  & 1.44 $\pm$ 0.06 \\ \cmidrule{2-7}
                    & \multirow{2}{*}{B-SPBN-FKDE-Simple} & 50         & 2.85 $\pm$ 0.89 &          -          & \hspace{1.65mm}1.47e+02 $\pm$ 34.84 &           -         \\
                    &                            & 100        & 1.86 $\pm$ 0.11 &          -          & 1.86e+01 $\pm$ 0.81  &           -         \\ \hline
\multirow{4}{*}{6}  & \multirow{2}{*}{B-SPBN-Simple}      & 50  & \textbf{2.12 $\pm$ 0.87} & 1.20 $\pm$ 0.14 & 3.10e+01 $\pm$ 2.80 & 1.30 $\pm$ 0.21\\
                    &                                     & 100 & 1.67 $\pm$ 0.26 & 1.23 $\pm$ 0.09 & 9.95e+00 $\pm$ 0.48 & 1.34 $\pm$ 0.09\\ \cmidrule{2-7}
                    & \multirow{2}{*}{B-SPBN-FKDE-Simple} & 50  & 3.84 $\pm$ 1.00 & - & \hspace{1.7mm}\textbf{1.56e+02 $\pm$ 12.25} & -\\  
                    &                                     & 100 & \textbf{3.12 $\pm$ 0.57} & - & \textbf{3.19e+01 $\pm$ 1.30}  & -\\ \hline
                    
\multirow{4}{*}{7}  & \multirow{2}{*}{B-SPBN-Simple}      & 50  & \textbf{2.46 $\pm$ 0.50} & \textbf{2.53 $\pm$ 0.14} & \hspace{1.7mm}\textbf{7.91e+01 $\pm$ 11.83} & 3.37 $\pm$ 0.52\\
                    &                                     & 100 & \textbf{2.23 $\pm$ 0.22} & \textbf{1.55 $\pm$ 0.46} & \textbf{2.74e+01 $\pm$ 3.41} & 2.09 $\pm$ 0.22\\ \cmidrule{2-7}
                    & \multirow{2}{*}{B-SPBN-FKDE-Simple} & 50  & 2.82 $\pm$ 0.23 & - & \hspace{1.7mm}\textbf{1.47e+02 $\pm$ 10.66} & -\\
                    &                                     & 100 & 2.05 $\pm$ 0.09 & - & 2.36e+01 $\pm$ 1.90   & -\\
\bottomrule
\end{tabular}
\caption{B-SPBN time ratios.}\label{tab:results_real}
\end{table}

Table \ref{tab:results_real} shows that all B-SPBNs, regardless of whether they use SBKDE or FKDE CPDs, were on average faster than the SPBN. The two best ratios of each model for \( M = 50 \) and \( M = 100 \) in each column are highlighted in bold. Many of these ratios correspond to datasets 2 and 3, particularly for B-SPBNs with multiple parent nodes, where HC ratios of 2 and test ratios of 10 are observed. The primary reason for this trend is the maximum number of parents allowed in the DAG, as these datasets contain fewer variables. Binning the data requires constructing \( n \)-dimensional tensors. By using sparse tensors, we optimize the memory usage and computation of kernel densities. However, as dimensionality increases, it becomes less likely that data points will be grouped into the same grid vector compared to lower-dimensional cases. Thus, SBKDE CPDs using the simple binning rule converge to the complexity of CKDE CPDs as the number of parents in a particular node grows.
This explanation aligns with the best ratios observed in the columns for a single parent node, where some of the top test ratios were achieved by both B-SPBN-Simple and B-SPBN-FKDE-Simple in dataset 7 instead of dataset 2. Particularly, the best result was achieved by B-SPBN-FKDE-Simple in dataset 6 with a test ratio of 156.

On the other hand, Figure \ref{fig:bergmann_real} illustrates that there are no statistically significant differences between SPBNs and B-SPBNs, regardless of whether they have one or multiple parent nodes. However, significant differences are observed when comparing these models to GBNs. Within the semiparametric group, the SPBN ranked the highest, and the B-SPBN-FKDE-Simple apparently outperformed (although the difference is not statistically significant) the B-SPBN-Simple. Nevertheless, all error and structural distance metrics indicated the opposite previously. A possible explanation concerns the data binning and FFT-based computation of the KDE, which could lead to higher log-likelihood estimations without translating into better structures, as evidenced by the synthetic experiments.

\section{Conclusion}
\label{sec:conclusion}
This paper has introduced the B-SPBNs, an improved SPBN that accelerates the estimation of CKDE CPDs in nonparametric distributions. The acceleration is performed by taking advantage of data binning properties. Thus, two new types of CPDs, the SBKDE CPDs and the FKDE CPDs, are defined in substitution of the CKDE CPDs. Both contributions are derived from a binned computation of the conventional KDE, named BKDE. In the SBKDE CPDs, the BKDE equation is modified to account for sparse tensors, reducing the computational cost and memory requirements associated with a higher number of variables. In contrast, FKDE CPDs are restricted to a low dimensionality to avoid the curse of dimensionality. Here, the summation in the BKDE model is transformed into a convolution-like equation that can be solved with the FFT, returning much faster results.

The experiments showed that SBKDE CPDs produce results comparable to CKDE CPDs, with execution times 2 times faster for the structure learning and up to 10 times faster for the log-likelihood estimation. FKDE CPDs exhibit higher error rates, but they offer significant speed advantages, as the structure of the SPBN can be obtained 2 to 4 times faster and the log-likelihood up to 156 times faster. Moreover, a Friedman test followed by a Bergmann-Hommel \textit{post-hoc} analysis showed no significant structural differences between the two. These advantages are particularly noticeable when the number of parent nodes is small, since FKDE CPDs are limited to a few parents, and the improvement of SBKDE CPDs over CKDE CPDs decreases as this number grows.  

Several aspects could benefit from further research, such as comparing the B-SPBNs with other advanced non-Bayesian network-based approaches (e.g., Gaussian mixture models or normalizing flows) to evaluate their competitiveness in a broader field, or adapting the FKDE CPDs to handle more accurately a higher number of variables. In addition, there are contributions in the literature that could be applied to allow for adaptive or automatic grid selection, which were not included in this paper. Finally, the speed improvements achieved by the B-SPBNs could benefit industrial applications, particularly in edge device deployments.

\section*{Declaration of Generative AI and AI-assisted technologies in the writing process}
During the preparation of this work the author(s) used GPT-4o mini in order to improve readability. After using this tool/service, the author(s) reviewed and edited the content as needed and take(s) full responsibility for the content of the publication.

\section*{Acknowledgments}
This work was partially supported by the Ministry of Science, Innovation and Universities under Project AEI/10.13039/501100011033-PID2022-139977NB-I00, Project TED2021-131310B-I00, Project PLEC2023-010252/MIG-20232016 and DIN2024-013310 (Doctorados Industriales). Also, by the Autonomous Region of Madrid under Project ELLIS Unit Madrid and TEC-2024/COM-89.

\section*{Data availability}
All our synthetic functions and research data are publicly available at: \url{https://github.com/rafasj13/BinnedSemiparametricBN}.  

\bibliographystyle{unsrt}


\appendix
\section{Synthetic SPBNs}
\label{apendix:synthetic_data}

Synthetic SPBN 1:
\begin{align}
    &f(a) \sim \mathcal{N}(\mu_A = 3, \sigma_A = 2) \nonumber \\
    &f(b|a) \sim \mathcal{N}(\mu_B = a \cdot 0.5, \sigma_B = 2) \nonumber \\
    &f(c|a) \sim 0.45 \cdot \mathcal{N}(\mu_{C_1} = a \cdot 0.5, \sigma_{C_1} = 1.5) + 0.55 \cdot \mathcal{N}(\mu_{C_2} = 5, \sigma_{C_2} = 1) \nonumber \\
    &f(d|b, c) \sim 0.5 \cdot \mathcal{N}(\mu_{D_1} = c \cdot b \cdot 0.5, \sigma_{D_1} = 1) + 0.5 \cdot \mathcal{N}(\mu_{D_2} = 3.5, \sigma_{D_2} = 1) \\
    &f(e|d, c) \sim 0.5 \cdot \mathcal{N}(\mu_{E_1} = d + c, \sigma_{E_1} = 1) + 0.5 \cdot \mathcal{N}(\mu_{E_2} = 2, \sigma_{E_2} = 1) \nonumber \\
    &f(f|e, d, a) \sim 0.5 \cdot \mathcal{N}(\mu_{F_1} = e + d, \sigma_{F_1} = 1) + 0.5 \cdot \mathcal{N}(\mu_{F_2} = 0.7\cdot a, \sigma_{F_2} = 0.5) \nonumber \\
    &f(g|c) \sim \mathcal{N}(\mu_{G} = c \cdot 0.3, \sigma_{G} = 2) \nonumber
\end{align}

Synthetic SPBN 2:
\begin{align}
    &f(a) \sim \mathcal{N}(\mu_A = 4, \sigma_A = 1.5) \nonumber \\
    &f(b|a) \sim 0.4 \cdot \mathcal{N}(\mu_{B_1} = a \cdot 1.2, \sigma_{B_1} = 1.1) + 0.6 \cdot \mathcal{N}(\mu_{B_2} = 1, \sigma_{B_2} = 1) \nonumber \\
    &f(c|a) \sim 0.5 \cdot \mathcal{N}(\mu_{C_1} = a + 1, \sigma_{C_1} = 1.2) + 0.5 \cdot \mathcal{N}(\mu_{C_2} = 1, \sigma_{C_2} = 1) \nonumber \\
    &f(d|a) \sim \mathcal{N}(\mu_D = a \cdot 0.8, \sigma_D = 1.3) \nonumber \\
    &f(e|c) \sim 0.6 \cdot \mathcal{N}(\mu_{E_1} = c \cdot 1.2, \sigma_{E_1} = 1.3) + 0.4 \cdot \mathcal{N}(\mu_{E_2} = -1, \sigma_{E_2} = 1.5) \nonumber \\
    &f(h|d) \sim 0.6 \cdot \mathcal{N}(\mu_{H_1} = d \cdot 2, \sigma_{H_1} = 1.2) + 0.4 \cdot \mathcal{N}(\mu_{H_2} = 0, \sigma_{H_2} = 1.8) \nonumber \\
    &f(i|b) \sim \mathcal{N}(\mu_I = b \cdot 0.6, \sigma_I = 2)  \\
    &f(j|e) \sim \mathcal{N}(\mu_J = e \cdot 0.7, \sigma_J = 1.7) \nonumber \\
    &f(f|c, h) \sim 0.5 \cdot \mathcal{N}(\mu_{F_1} = c \cdot 1.1 + h, \sigma_{F_1} = 1) + 0.5 \cdot \mathcal{N}(\mu_{F_2} = 15, \sigma_{F_2} = 1.2) \nonumber \\
    &f(g|d, j) \sim 0.5 \cdot \mathcal{N}(\mu_{G_1} = d \cdot 0.8 + j, \sigma_{G_1} = 1) + 0.5 \cdot \mathcal{N}(\mu_{G_2} = 0, \sigma_{G_2} = 1) \nonumber \\
    &f(k|f) \sim \mathcal{N}(\mu_K = f \cdot 0.3, \sigma_K = 2) \nonumber \\
    &f(l|a, c, f, h, d) \sim 0.5 \cdot \mathcal{N}(\mu_{L_1} = a + c + f, \sigma_{L_1} = 1) + 0.5 \cdot \mathcal{N}(\mu_{L_2} = h \cdot 0.6 + d, \sigma_{L_2} = 1.5) \nonumber \\
    &f(m|b, e, g, j) \sim 0.4 \cdot \mathcal{N}(\mu_{M_1} = b + e + g, \sigma_{M_1} = 1.2) + 0.6 \cdot \mathcal{N}(\mu_{M_2} = j \cdot 0.7, \sigma_{M_2} = 1.3) \nonumber 
\end{align}

Synthetic SPBN 3:
\begin{align}
    &f(a) \sim 0.5 \cdot \mathcal{N}(\mu_{A_1} = 4, \sigma_{A_1} = 2) + 0.5 \cdot \mathcal{N}(\mu_{A_2} = 1, \sigma_{A_2} = 1) \nonumber \\ \nonumber
    &f(b|a) \sim \mathcal{N}(\mu_B = a \cdot 0.5, \sigma_B = 2) \\ \nonumber
    &f(c|b) \sim \mathcal{N}(\mu_C = b \cdot 2, \sigma_C = 1.5) \\ \nonumber
    &f(d|b) \sim 0.5 \cdot \mathcal{N}(\mu_{D_1} = b - 1, \sigma_{D_1} = 1) + 0.5 \cdot \mathcal{N}(\mu_{D_2} = 10, \sigma_{D_2} = 1.5) \\ 
    &f(e|d) \sim 0.5 \cdot \mathcal{N}(\mu_{E_1} = d \cdot 2, \sigma_{E_1} = 1.5) + 0.5 \cdot \mathcal{N}(\mu_{E_2} = 3, \sigma_{E_2} = 1) \\ \nonumber
    &f(f|d) \sim 0.6 \cdot \mathcal{N}(\mu_{F_1} = d \cdot 1.5, \sigma_{F_1} = 1.5) + 0.4 \cdot \mathcal{N}(\mu_{F_2} = 0, \sigma_{F_2} = 1) \\ \nonumber
    &f(g|c) \sim \mathcal{N}(\mu_G = c \cdot 0.3 + 5, \sigma_G = 1) \\ \nonumber
   &f(h|c) \sim 0.5 \cdot \mathcal{N}(\mu_{H_1} = c \cdot 0.5, \sigma_{H_1} = 1) + 0.5 \cdot \mathcal{N}(\mu_{H_2} = 10, \sigma_{H_2} = 1) 
\end{align}

Synthetic SPBN 4:
\begin{align}
    &f(a) \sim \mathcal{N}(\mu_A = 5, \sigma_A = 2) \nonumber \\
    &f(b|a) \sim \mathcal{N}(\mu_B = a +2, \sigma_B = 1.5) \nonumber \\
    &f(c|a) \sim 0.4 \cdot \mathcal{N}(\mu_{C_1} = a + 2, \sigma_{C_1} = 1) + 0.6 \cdot \mathcal{N}(\mu_{C_2} = 1, \sigma_{C_2} = 1.5) \nonumber \\
    &f(d|b) \sim 0.5 \cdot \mathcal{N}(\mu_{D_1} = b \cdot 0.8, \sigma_{D_1} = 1.5) + 0.5 \cdot \mathcal{N}(\mu_{D_2} = 15, \sigma_{D_2} = 1.5) \nonumber \\
    &f(e|c) \sim \mathcal{N}(\mu_E = c \cdot 0.7, \sigma_E = 2) \nonumber \\
    &f(f|c) \sim 0.5 \cdot \mathcal{N}(\mu_{F_1} = c \cdot 1.2, \sigma_{F_1} = 1.5) + 0.5 \cdot \mathcal{N}(\mu_{F_2} = -3, \sigma_{F_2} = 1) \nonumber \\
    &f(g|d) \sim 0.6 \cdot \mathcal{N}(\mu_{G_1} = d + 4, \sigma_{G_1} = 1) + 0.4 \cdot \mathcal{N}(\mu_{G_2} = 8, \sigma_{G_2} = 1.5) \nonumber \\
    &f(h|d) \sim \mathcal{N}(\mu_H = d \cdot 0.4, \sigma_H = 2)  \\
    &f(k|d) \sim \mathcal{N}(\mu_K = d \cdot 0.5, \sigma_K = 2.5) \nonumber \\
    &f(i|e) \sim 0.55 \cdot \mathcal{N}(\mu_{I_1} = e \cdot 1.3, \sigma_{I_1} = 2) + 0.45 \cdot \mathcal{N}(\mu_{I_2} = 0, \sigma_{I_2} = 1) \nonumber \\
    &f(j|e) \sim \mathcal{N}(\mu_J = e \cdot 0.5, \sigma_J = 2) \nonumber \\
    &f(o|f) \sim 0.3 \cdot \mathcal{N}(\mu_{O_1} = f + 1, \sigma_{O_1} = 1.4) + 0.7 \cdot \mathcal{N}(\mu_{O_2} = -2, \sigma_{O_2} = 0.7) \nonumber \\
    &f(m|j) \sim 0.6 \cdot \mathcal{N}(\mu_{M_1} = j \cdot 1.5, \sigma_{M_1} = 1) + 0.4 \cdot \mathcal{N}(\mu_{M_2} = 7, \sigma_{M_2} = 1.5) \nonumber \\
    &f(n|j) \sim 0.4 \cdot \mathcal{N}(\mu_{N_1} = j \cdot 1.1, \sigma_{N_1} = 1.2) + 0.6 \cdot \mathcal{N}(\mu_{N_2} = -1, \sigma_{N_2} = 1.3) \nonumber \\
    &f(l|h) \sim 0.5 \cdot \mathcal{N}(\mu_{L_1} = h \cdot 0.3, \sigma_{L_1} = 1.1) + 0.5 \cdot \mathcal{N}(\mu_{L_2} = 5, \sigma_{L_2} = 1.4) \nonumber
\end{align}

Synthetic SPBN 5:
\begin{align}
    &f(a) \sim \text{Exp}(\lambda_A = 1) \nonumber \\
    &f(b|a) \sim \text{Exp}(\lambda_B = \tfrac{1}{a}) \nonumber \\
    &f(c|a) \sim \text{Exp}(\lambda_C = \tfrac{1}{2a}) \nonumber \\
    &f(d|b, c) \sim \text{Exp}(\lambda_D = \tfrac{1}{b \cdot c}) \\
    &f(e|d, c) \sim \text{Exp}(\lambda_E = \tfrac{1}{d \cdot c}) \nonumber \\
    &f(f|a, d, e) \sim \text{Exp}(\lambda_F = \tfrac{1}{a + 2d + e}) \nonumber \\
    &f(g|c) \sim \text{Exp}(\lambda_G = \tfrac{1}{c}) \nonumber
\end{align}

Synthetic SPBN 6:
\begin{align}
    &f(a) \sim \mathrm{Gam}(k_A = 2, \theta_A = 1) \nonumber \\
    &f(b|a) \sim \mathrm{Gam}(k_B = a, \theta_B = 1) \nonumber \\
    &f(c|b) \sim \mathrm{Gam}(k_C = b, \theta_C = 1) \nonumber \\
    &f(d|b) \sim \mathrm{Gam}(k_D = b, \theta_D = 1) \\
    &f(e|d) \sim \mathrm{Gam}(k_E = d, \theta_E = 1) \nonumber \\
    &f(f|d) \sim \mathrm{Gam}(k_F = d, \theta_F = 1) \nonumber \\
    &f(g|c) \sim \mathrm{Gam}(k_G = c, \theta_G = 1) \nonumber \\
    &f(h|c) \sim \mathrm{Gam}(k_H = c, \theta_H = 1) \nonumber
\end{align}

Synthetic SPBN 7:
\begin{align}
    &f(a) \sim \text{Beta}(\alpha_A = 2, \beta_A = 8) \nonumber \\
    &f(b|a) \sim \text{Beta}(\alpha_B = a, \beta_B = 2) \nonumber \\
    &f(c|a) \sim \text{Beta}(\alpha_C = a, \beta_C = 4) \nonumber \\
    &f(d|b, c) \sim \text{Beta}(\alpha_D = b, \beta_D = c) \\
    &f(e|d, c) \sim \text{Beta}(\alpha_E = d, \beta_E = c) \nonumber \\
    &f(f|a, d, e) \sim \text{Beta}(\alpha_F = a + 2d, \beta_F = e) \nonumber \\
    &f(g|c) \sim \text{Beta}(\alpha_G = 1, \beta_G = c) \nonumber
\end{align}

Synthetic SPBN 8:
\begin{align}
    &f(a) \sim \mathrm{Lap}(\mu_A = 5, b_A = 2) \nonumber \\
    &f(b|a) \sim \mathrm{Lap}(\mu_B = a, b_B = 2) \nonumber \\
    &f(c|b) \sim \mathrm{Lap}(\mu_C = b, b_C = 2) \nonumber \\
    &f(d|b) \sim \mathrm{Lap}(\mu_D = b, b_D = 2) \\
    &f(e|d) \sim \mathrm{Lap}(\mu_E = d, b_E = 2) \nonumber \\
    &f(f|d) \sim \mathrm{Lap}(\mu_F = d, b_F = 2) \nonumber \\
    &f(g|c) \sim \mathrm{Lap}(\mu_G = c, b_G = 2) \nonumber \\
    &f(h|c) \sim \mathrm{Lap}(\mu_H = c, b_H = 2) \nonumber
\end{align}

\end{document}